\documentclass{bmvc2k}

\usepackage{epsfig}
\usepackage{graphicx}
\usepackage{amsmath}
\usepackage{amssymb}

\usepackage{times}

\usepackage{soul}
\usepackage{url}
\usepackage[utf8]{inputenc}
\usepackage[small]{caption}
\usepackage{booktabs}
\usepackage{threeparttable}
\usepackage{adjustbox}
\usepackage{makecell}
\usepackage{multicol}
\usepackage{multirow}
\usepackage{xcolor}         

\title{Lightweight HDR Camera ISP for Robust Perception in Dynamic Illumination Conditions via Fourier Adversarial Networks}

\addauthor{Pranjay Shyam}{pranjayshyam@kaist.ac.kr}{1}
\addauthor{Sandeep Singh Sengar}{sengar@di.ku.dk}{2}
\addauthor{Kuk-Jin Yoon}{kjyoon@kaist.ac.kr}{1}
\addauthor{Kyung-Soo Kim}{kyungsookim@kaist.ac.kr}{1}

\addinstitution{
 Korea Advanced Institute of Science and Technology (KAIST) \\
 Daejeon, Republic of Korea
}
\addinstitution{
 University of Copenhagen\\
 Copenhagen, Denmark
}

\runninghead{Shyam, Sengar, Yoon, Kim}{Lightweight HDR Camera ISP}

\begin{document}

\maketitle

\begin{abstract}
\noindent
The limited dynamic range of commercial compact camera sensors results in an inaccurate representation of scenes with varying illumination conditions, adversely affecting image quality and subsequently limiting the performance of underlying image processing algorithms. Current state-of-the-art (SoTA) convolutional neural networks (CNN) are developed as post-processing techniques to independently recover under-/over-exposed images. However, when applied to images containing real-world degradations such as glare, high-beam, color bleeding with varying noise intensity, these algorithms amplify the degradations, further degrading image quality. We propose a lightweight two-stage image enhancement algorithm sequentially balancing illumination and noise removal using frequency priors for structural guidance to overcome these limitations. Furthermore, to ensure realistic image quality, we leverage the relationship between frequency and spatial domain properties of an image and propose a Fourier spectrum-based adversarial framework (AFNet) for consistent image enhancement under varying illumination conditions. While current formulations of image enhancement are envisioned as post-processing techniques, we examine if such an algorithm could be extended to integrate the functionality of the Image Signal Processing (ISP) pipeline within the camera sensor benefiting from RAW sensor data and lightweight CNN architecture. Based on quantitative and qualitative evaluations, we also examine the practicality and effects of image enhancement techniques on the performance of common perception tasks such as object detection and semantic segmentation in varying illumination conditions.
\end{abstract}

\vspace{-5mm}
\section{Introduction}
\vspace{-2mm}
Images captured in dynamic illumination conditions can have underexposed or overexposed regions or a combination of both. The underexposed regions are susceptible to noise, and overexposed regions obscure textural information of surrounding features. This deteriorates the performance of underlying high-level perception tasks such as feature matching \cite{shyam2020retaining}, lane detection \cite{liu2020lane}, object detection \cite{michaelis2019benchmarking}, and semantic segmentation \cite{sun2019see}. While hardware modifications or software adjustments can be used for increasing light received by a camera sensor when capturing a scene, these approaches introduce additional noise and artifacts such as motion blur (increasing exposure time), losing the depth of field (increasing aperture of the appropriate lens), and non-uniform lightening (using additional light source). Hence focus shifts towards software-based image enhancement as a post-processing technique to enhance image quality while maintaining image sharpness and color balance.

\begin{figure}
\centering
\begin{adjustbox}{width=0.99\columnwidth}
\renewcommand{\tabcolsep}{3pt} 
\begin{tabular}{c c}
\includegraphics[width=0.45\textwidth, height=5.5cm, trim={0 2.3cm 0 0}, clip]{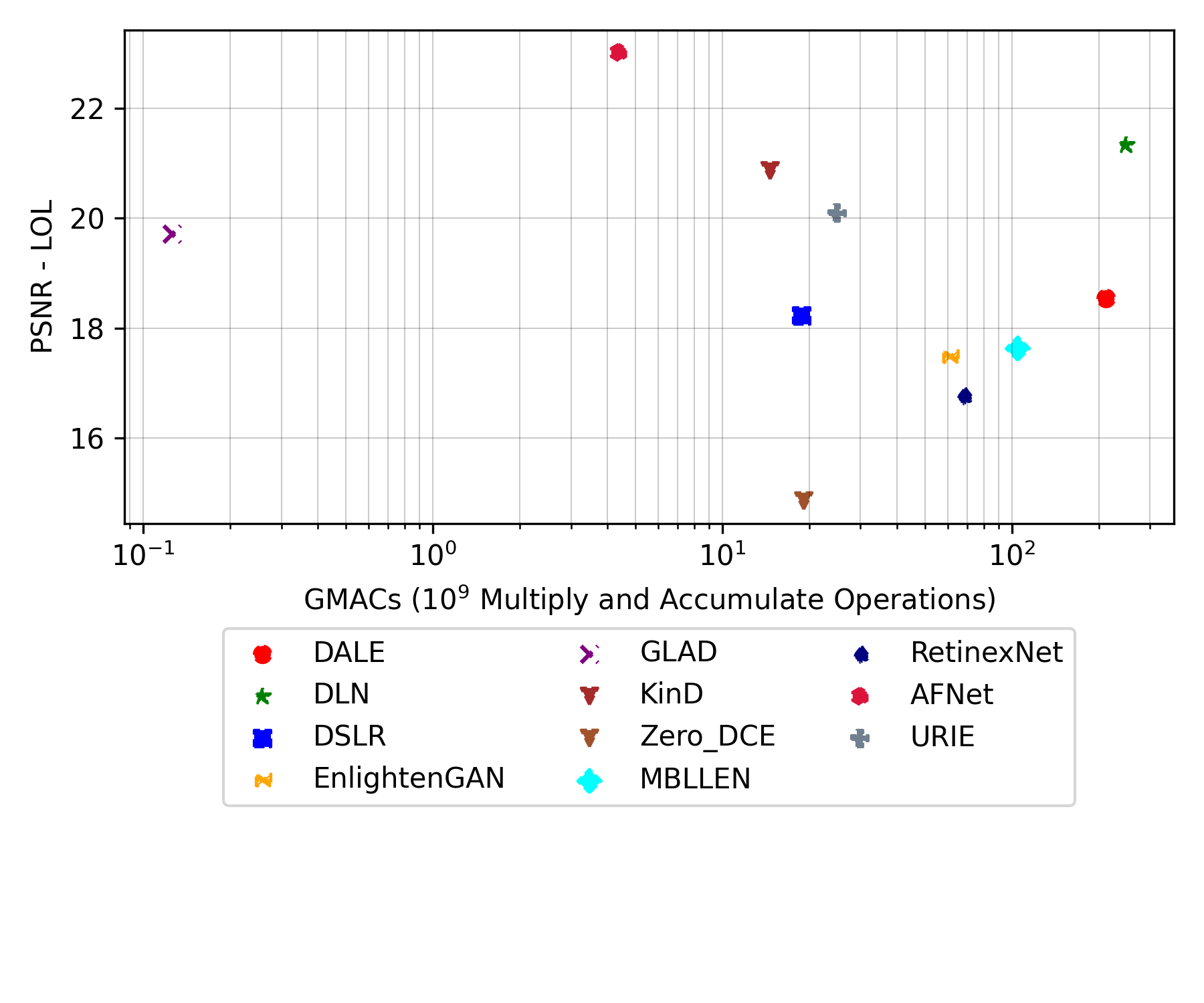} & 
\includegraphics[width=0.55\textwidth, height=5.5cm]{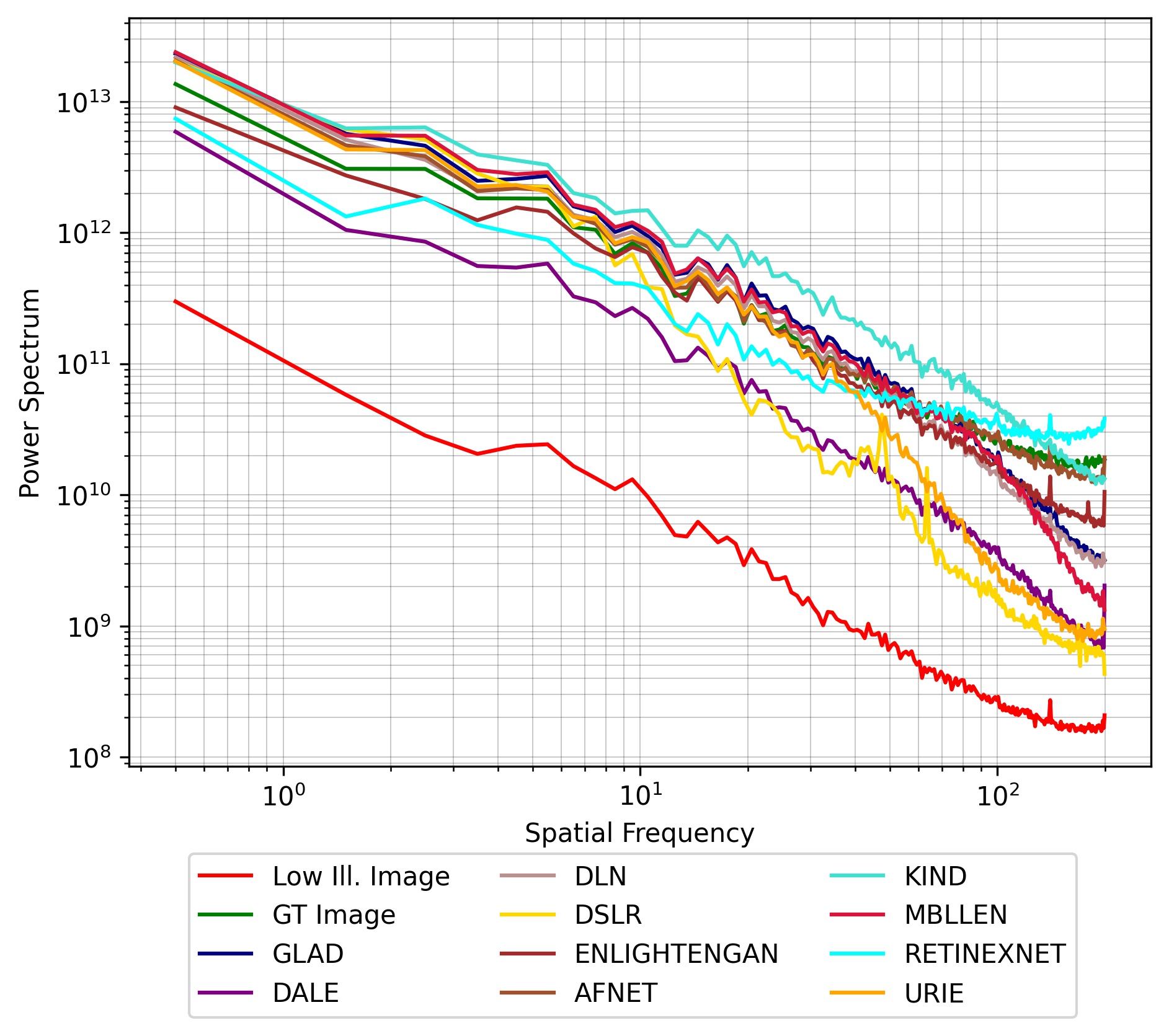} \\ \vspace{1pt}
\end{tabular}
\end{adjustbox}
\caption{Performance landscape (GMACs vs PSNR) of different SoTA Image Enhancement Algorithms on sRGB Images from the LOL \cite{wei2018deep} dataset (left) and corresponding Power Spectral Density curves of enhanced images (right).}
\label{fig_1}
\vspace{-5mm}
\end{figure}

Current SoTA algorithms leverage CNNs and define different functional configurations focusing on CNN architectures \cite{lore2017llnet, zhang2019kindling, zhang2020attention} or optimization formulation \cite{jiang2019enlightengan, zhang2020self} to obtain a well-illuminated image in low light or high illumination conditions. However, illumination settings confine the performance of these methods; hence they perform well only in the conditions wherein the complete image has similar illumination conditions. This assumption is rarely fulfilled in real scenarios, resulting in increased pixel noise, color bleeding, and pixelations, reducing image quality when using these algorithms on natural images containing local illumination sources. This is extremely detrimental in scenarios wherein these enhanced images are used as inputs for performing high-level vision tasks such as object detection, semantic segmentation, etc., as it degrades the performance of SoTA algorithms (See the supplementary).

To circumvent these limitations of SoTA low-light image enhancement (LLIE) algorithms, we propose a two-stage enhancement architecture wherein the first stage focuses on coarsely balancing illumination and the second stage focuses on noise and artifact removal to reconstruct a well-illuminated color-balanced image. Furthermore, to improve feature quality without increasing computations, we propose a compact multi-scale feature extraction mechanism that splits a given feature map along channel dimension and subsequently uses a convolutional filter with different kernel sizes. These features are then aggregated after being scaled using a channel attention mechanism that encourages relevant features while suppressing irrelevant features, allowing us to obtain features across diverse receptive fields used to balance the illumination of the image. Subsequently, the secondary network is used to recover the regions affected by noise and artifacts to ensure textural and structural fidelity within enhanced images. This two-stage approach reduces the network size while ensuring SoTA performance, saving on inference time and memory requirement.

\begin{figure}[!t]
\centering
\scriptsize
\renewcommand{\tabcolsep}{3pt} 
\begin{minipage}{0.6\linewidth}\centering
\renewcommand{\tabcolsep}{1pt} 
\renewcommand{\arraystretch}{1} 
\begin{tabular}{cccc}
\includegraphics[width=0.225\columnwidth, height=2cm]{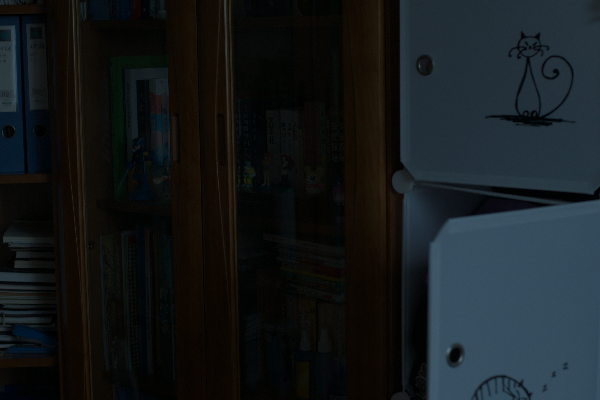} &
\includegraphics[width=0.225\columnwidth, height=2cm]{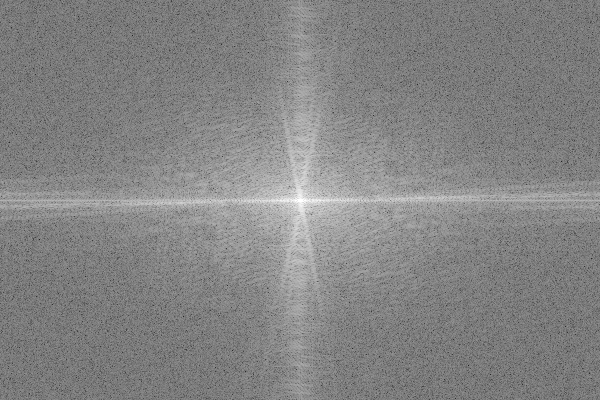} &
\includegraphics[width=0.225\columnwidth, height=2cm]{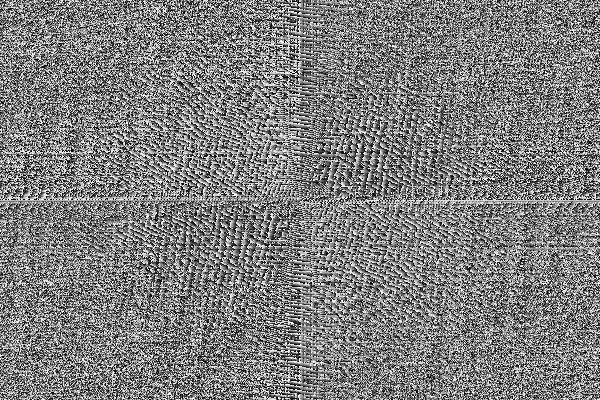} & 
\includegraphics[width=0.225\columnwidth, height=2cm, trim={2cm 2cm 2cm 2cm}, clip]{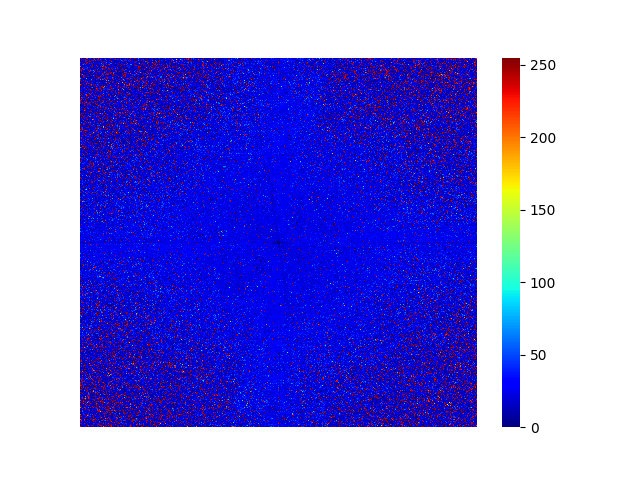} \\
(a) LL Image & (b) Magnitude & (c) Phase & (d) $\delta$ Magnitude \\
\includegraphics[width=0.225\columnwidth, height=2cm]{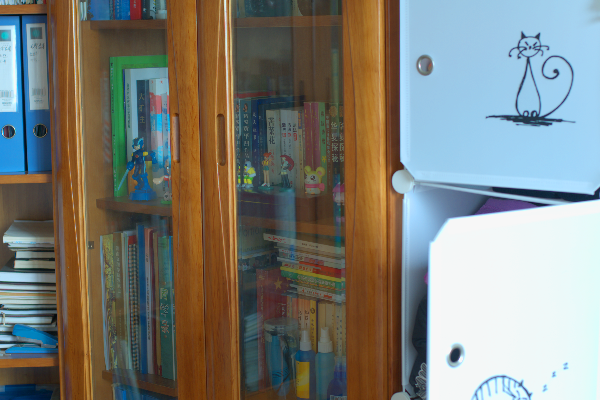} &
\includegraphics[width=0.225\columnwidth, height=2cm]{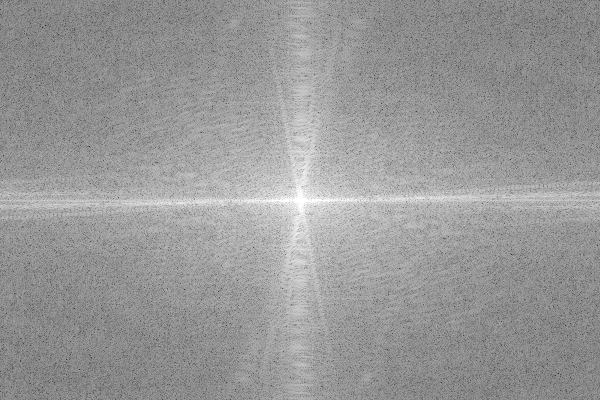} &
\includegraphics[width=0.225\columnwidth, height=2cm]{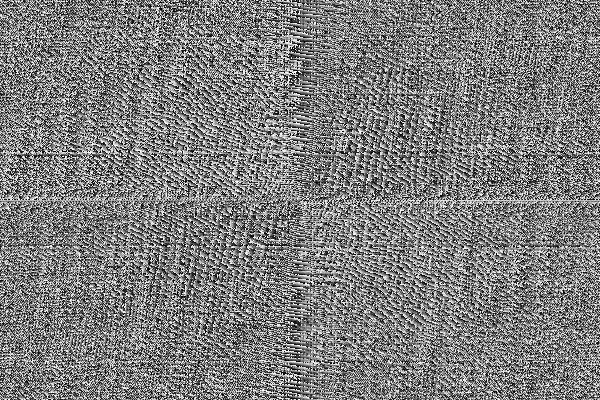} & 
\includegraphics[width=0.225\columnwidth, height=2cm, trim={2cm 2cm 2cm 2cm}, clip]{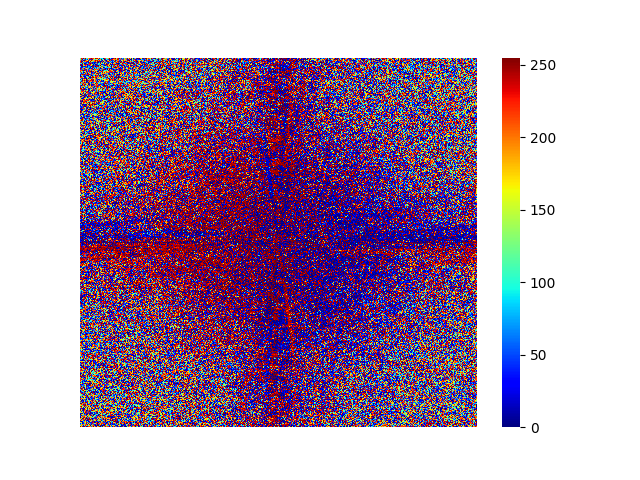} \\
(e) GT & (f) Magnitude & (g) Phase & (h) $\delta$ Phase \\ \\
\end{tabular}
\caption{Disentangling (a) low light and (e) its corresponding ground truth into (b, f) magnitude and (c, g) phase components using fast Fourier transform with difference heatmap of magnitude and phase (d, h). In the heatmap, red highlights maximum error whereas blue represents minimum error.}
\label{fig_2}
\end{minipage}
\quad
\begin{minipage}{0.35\linewidth}
\includegraphics[width=\columnwidth, height=5cm]{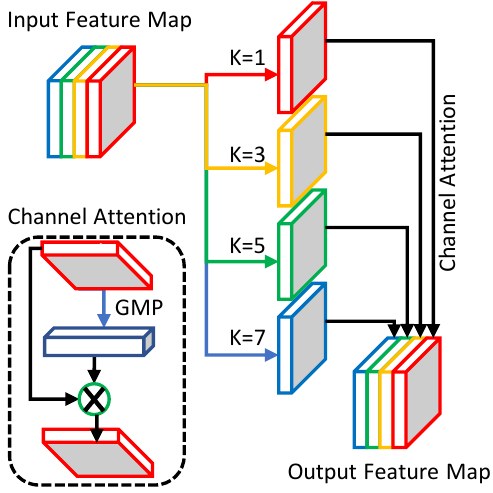} \\
\caption{Overview of the proposed compact multiscale-feature extraction and aggregation.}
\label{fig_3}
\end{minipage}
\vspace{-5mm}
\end{figure}

Furthermore, upon a closer inspection of power spectral density of images enhanced by SoTA algorithms, we observe poor performance at high frequencies that capture edge information; thus, we propose utilizing the frequency domain information to ensure consistent enhancement under diverse conditions by leveraging the duality between frequency and spatial domain characteristics of an image. Specifically, point-wise modifications in the frequency domain result in global modifications across spatial domains in an image. In addition, visual examination (Fig. \ref{fig_2}) of frequency domain information, i.e., magnitude and phase components generated using Fast Fourier Transform (FFT), reveals multiple attributes that can be leveraged to ensure image enhancement. Some notable attributes include the presence of high textural information within a well-lit image (Fig. \ref{fig_2}(e)) that are centered in the magnitude spectrum. As low light image doesn't capture detailed textural information, the intensity of magnitude spectrum (Fig. \ref{fig_2}(b)) is attenuated. This observation can be extended for edges present in an image. 
While the magnitude spectrum can be interpreted as 'how much' of frequencies are present in an image, an equally important phase component (Fig. \ref{fig_2}(c, f)) determines 'where' those frequencies are present in the image. This motivates us to construct an adversarial network that utilizes both the magnitude and phase components of an image to determine whether it is real/fake. Such a binary CNN would leverage the complete spectral properties of how much and where certain frequencies are present and thus result in enhanced images closely resembling the ground truth. 

While the performance of image enhancement algorithms has improved lately, utilization of camera-ISP (comprising of multiple handcrafted task-specific algorithms to convert raw color filter array (CFA) data to standard RGB (sRGB) image) introduces additional non-linearities capping the peak performance of SoTA algorithms. However, due to the proprietary nature of camera-ISP, the implementation of enhancement algorithms on RAW sensor data is not well studied. Recently different works such as SID \cite{chen2018learning}, and Five5K \cite{fivek} have collected RAW image pairs using different exposure times to construct paired images that could be used for training end-to-end image enhancement algorithms instead of the current post-processing formulation. However, as these datasets capture high illumination conditions by increasing the exposure times, they do not contain dynamic scenarios such as glares, color bleeding, high beam, etc., thus representing conditions that are easily violated in real-life deployment. Nevertheless, they could be used for training an end-to-end CNN with additional augmentations as paired-dataset capturing diverse illumination conditions and storing it in RAW format is impractical. Instead, we construct a test-set that represents varying illumination conditions under diverse conditions by capturing images using a dashboard camera mounted on a consumer vehicle, allowing us to examine the efficacy of algorithms on real-world deployment. We summarize our contributions as,

\vspace{-2mm}
\begin{itemize}
\item We propose a two-stage CNN architecture for performing illumination balancing and image restoration that works with both sRGB and RAW images. \vspace{-2mm}
\item We combine channel split mechanism with multiscale convolutions to enhance the receptive field and increase feature information without a substantial increase in computations. \vspace{-2mm}
\item To ensure the presence of high-frequency components within enhanced images, we propose using frequency information within an adversarial learning mechanism. \vspace{-2mm}
\item As paired training datasets cannot represent dynamic conditions, we construct an unpaired test-set by collecting RAW and sRGB images under dynamic illumination conditions using a personal vehicle as data capturing setup. \vspace{-2mm}
\item We demonstrate varying illumination conditions to adversely affect the performance of object detection algorithms and improve it by enhancing image quality using the proposed approach. \vspace{-2mm}
\end{itemize}

\vspace{-5mm}
\section{Related Works}
\vspace{-3mm}
\subsection{Image Enhancement} 
\vspace{-2mm}
Early CNN-based approach, LLNet \cite{lore2017llnet}, proposed an autoencoder formulation for performing contrast enhancement while simultaneously suppressing noise. Subsequent works rely on Retinex Theory coupled with additional priors such as structure aware loss in RetinexNet \cite{wei2018deep}, reflectance restoration in KinD \cite{zhang2019kindling}, and attention mechanism \cite{zhang2020attention} to improve LLIE performance. When applying these algorithms on images comprising both well and poorly-lit regions, they distort the regions that do not require any enhancement. To overcome such situations, DALE \cite{kwon2020dale} proposed a two-stage approach of first identifying dark regions using a visual attention module and then enhancing the brightness of these regions. These methods perform LLIE on images of reduced spatial resolution resulting in inaccurate spatial enhancement. MIRNet \cite{Zamir2020MIRNet} was proposed to maintain a semantically and spatially accurate enhancement network using multi-resolution convolution and attention mechanisms. As these methods require paired training samples, constructing a training dataset is extremely time-consuming. EnlightenGAN \cite{jiang2019enlightengan} utilized a generative adversarial framework for constructing low light images and subsequently using it to train an underlying LLIE algorithm, whereas \cite{zhang2020self} relying on self-supervised learning to formulate a retinex model optimized using maximum entropy. 

Recently frequency priors have been explored to restore images with MWCNN \cite{Liu_2018_CVPR_Workshops} using wavelet transforms to perform tasks such as super-resolution, denoising, and JPEG artifact removal. \cite{Xu_2020_CVPR} highlighted that detecting and removing noise is much easier from low-frequency components and thus proposed a two-stage network that decomposes an image into low and high frequency, recovers low-frequency components, and enhances high-frequency details. In addition, DIDH \cite{shyam2020towards} proposed an adversarial framework utilizing low and high-frequency prior-based discriminators for domain invariant dehazing. While these works highlight information represented within the frequency domain and devise different strategies for exploiting it for restoration tasks, we use an adversarial Fourier network to leverage its duality property with an image for performing region-sensitive image enhancement. As sRGB images are widely available and used for conducting research for high-level perception tasks, \cite{Exdark} constructed a dataset to demonstrate that low illumination conditions obscure information contained within an image, resulting in a performance drop of SoTA object detectors. Theoretically, the performance can be retained or improved if the image is processed using an ideal enhancement algorithm. However, from our experiments, we demonstrate that current methods result in increased pixelation and noise that adversely affect performance. Hence an ideal image enhancement algorithm is still missing. 

\vspace{-3mm}
\subsection{End-to-End Camera ISP} 
\vspace{-2mm}
Traditional ISP comprises multiple low-level tasks such as white balancing, demosaicing of CFA data, denoising, high dynamic range compression, black pixel removal, contrast enhancement, tone mapping, super-resolution, etc. The order of application and additional algorithms are unique to sensor manufacturers and inaccessible in most cases with extensive studies being conducted for independently performing these low-level tasks with state-of-the-art (SoTA) performance achieved using CNNs. Furthermore, due to the electronic nature of the camera sensor, it is prone to various noise from various sources such as photon noise, quantization noise, and digital noise \cite{guan2019node, wei2020physics}. This motivated different works such as \cite{remez2017deep, abdelhamed2020ntire, dabov2007image} to focus on removing noise to improve the signal-to-noise ratio, which has a more prominent effect on images captured in low light conditions \cite{chen2018learning, wei2020physics} due to low pixel intensities. \cite{lee2016joint} observed superior performance of dehazing algorithm with reduced artifacts when the RAW image is used instead of sRGB image. Encouraged by the success of individual CNNs on low-level tasks, \cite{kokkinos2019iterative} proposed a solution to jointly perform denoising and demosaicing using a residual connection to improve feature flow and better leverage image structure. \cite{qian2019trinity} extended this approach by integrating the task of super-resolution and jointly optimizing the underlying CNN to obtain high-quality RGB images. To further improve the quality of sRGB images \cite{schwartz2018deepisp, liang2019cameranet} proposed a two-stage framework for sequentially restoring and enhancing an image. Lately \cite{ignatov2020replacing} proposed an end-to-end framework for mapping a RAW demosaiced image captured via smartphone camera into sRGB space while simultaneously enhancing it to match the quality with a DSLR camera. Similar to these approaches, we perform end-to-end RAW-to-sRGB image conversion while removing illumination inconsistencies to obtain a balanced image using a demosaiced image as input and improving performance using frequency priors while achieving real-time performance.

\vspace{-6mm}
\section{Methodology}
\vspace{-3mm}

\subsection{Problem Formulation}
\vspace{-3mm}
Functioning of current SoTA image enhancement algorithms is limited to either low illumination or high contrast conditions while being capped by non-linearities arising from camera-ISP. This increases the computational cost of current SoTA, making them unviable for real scenarios wherein such image enhancement mechanisms can improve the performance of different perception tasks. As frequency spectrum is beneficial in ascertaining the limitations of current SoTA algorithms, we integrate such information in the CNN architecture and optimization cycle to ensure textural and structural consistency within enhanced images without noise or unwanted artifacts, thereby providing high dynamic range without relying upon multiple images.

\begin{figure}[!t]
\centering
\includegraphics[width=0.99\textwidth, height=5cm]{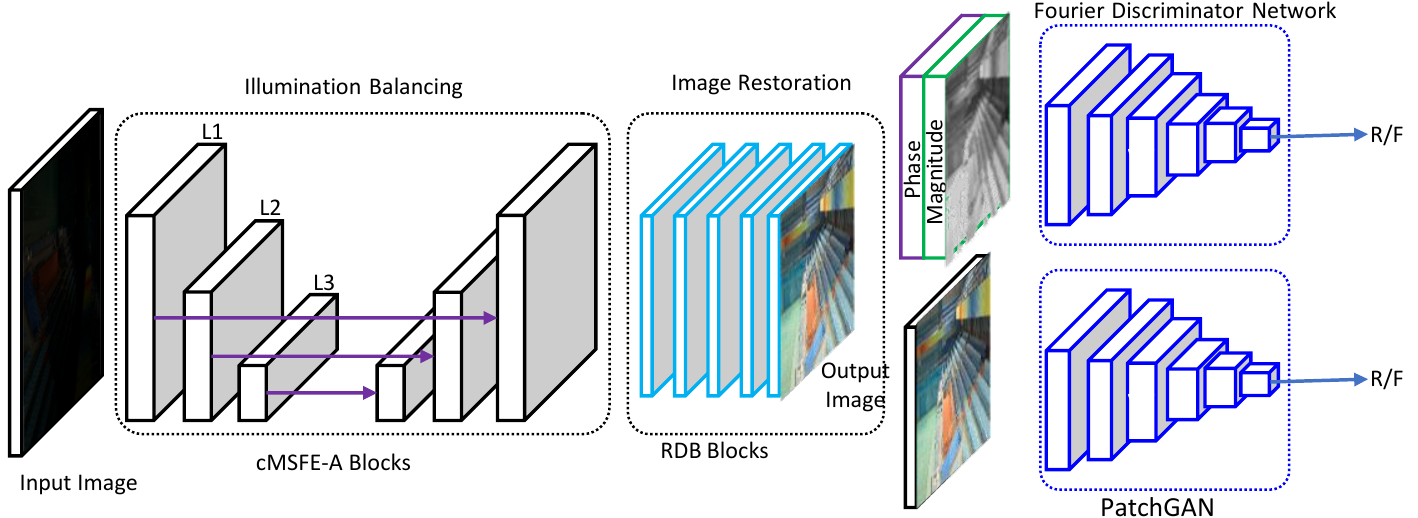}
\vspace{2mm}
\caption{Overview of the proposed image enhancement framework}
\vspace{-5mm}
\label{fig_4}
\end{figure} 

\vspace{-3mm}
\subsection{Network Architecture}
\vspace{-2mm}
Our baseline architecture comprises a two-stage process wherein the first stage enhances illumination, and the second stage removes any artifacts (Fig. \ref{fig_4}). We utilize different techniques introduced in the literature to develop efficient modules that provide high feature quality with reduced computational resources. 

\noindent
\textbf{Illumination Balancing} - As different regions can be affected by various illumination sources, to ensure consistent illumination, we use a UNet \cite{ronneberger2015u} style encoder-decoder framework that allows convolutional kernels access to complete image at the encoder end. To avoid excessive computations arising from extracting features across multiple scales (1/2, 1/4, 1/8, 1/16, 1/32), inspired by performance gains achieved by increasing receptive field size of convolutional layers,  we instead focus on achieving compact multi-scale feature extraction mechanism and thus extract features from 3 scales, i.e., 1/2, 1/8 and 1/32. Using multi-scale convolutions increases computational cost, hence to reduce the computational cost, different techniques such as channel shuffling, \cite{zhang2018shufflenet}, squeeze-excitation \cite{hu2018squeeze}, inception modules \cite{szegedy2015going}, 1x1 point convolutions \cite{lin2013network} etc., were proposed. In this paper, we combine these techniques to obtain a computationally efficient multi-scale feature extraction mechanism (Fig. \ref{fig_3}) and propose a channel split mechanism that divides a feature map across channel dimensions into multiple parts ($P$) of equal size (implying number of channels $(C)$ should be divisible by a split factor $S$). These parts are then used as inputs for convolutional kernels of different filter sizes to obtain features across a wider receptive field. To ensure relevant scale-specific features are amplified, we integrate a channel attention mechanism to features from each scale which are subsequently aggregated.

\noindent
\textbf{Image Restoration - } Upon enhancing illumination within the image, different artifacts and noises can be effectively restored. Furthermore, to ensure the presence of structural and textural details, we concatenate the original input image along with enhanced image and use residual dense blocks (RDBs) \cite{zhang2018residual} (that combine local and global features) to ensure similarity with ground truth. While RDBs are usually used on down-scaled features, we argue that subsequent upsampling of these features would reduce image quality resulting in losing details. Contrarily using them on complete images avoids these issues. 

\vspace{-4mm}
\subsection{Fourier Adversarial Network} \label{loss_formulation}
\vspace{-2mm}
As there is a perceivable difference between Fourier transforms of low light and corresponding ground truth images (Fig. \ref{fig_2}), frequency domain information (extracted using Fast Fourier Transform (FFT)) can be used to improve image quality by incorporating it within the optimization cycle. Furthermore, as low and high frequencies can be used concurrently to capture structural information better, using a complete Fourier spectrum can ensure structural consistency within the enhanced image. While pixel-based losses could be used to ensure FFT of enhanced and ground truth images are similar, they fail to capture inter-pixel relationships across neighboring pixels. Instead, we propose to use a CNN-based binary loss to determine whether a given image is real or fake based on the Fourier spectrum of its grayscale version. Furthermore, since the magnitude spectrum of an image has higher intensity around zero frequency, we normalize it before concatenating it with an input image that is then passed to the discriminator for binary classification. As we use a Fourier-based adversarial network to identify real/fake images using structural details, we require another adversarial network to ensure equal balance towards textural details. Thus we use commonly used PatchGAN \cite{pix2pix2017} for this purpose.  

In summary, the complete framework comprises a two-stage enhancement network that acts as a generator (G) with two discriminators focusing on structural (D1) and textural (D2) details to determine genuinity of a given image. For optimizing the complete framework, a combination of pixel (L1), structural (MS-SSIM \cite{1292216}), and feature-based (Supervised Contrastive Adversarial Loss) loss functions along with adversarial losses (following LSGAN \cite{mao2017least}) are used resulting in the following optimization objective for learnable parameters within generator ($\theta_G$) and discriminators ($\theta_{D1}, \theta_{D2}$),

\vspace{-4mm}
\begin{equation}
\begin{split}
\underset{\theta_G}{\min} & \ \underset{\theta_{D1}, \theta_{D2}}{\max} \quad \lambda_{L1}\mathbb{L}_{L1}(\theta_G) + \lambda_{MS-SSIM}\mathbb{L}_{MS-SSIM}(\theta_G) \\ + \ & \lambda_{SCAL}\mathbb{L}_{SCAL}(\theta_G) + \lambda_{P-ADV}\mathbb{L}_{P-ADV}(\theta_G, \theta_{D1}) + \lambda_{F-ADV}\mathbb{L}_{F-ADV}(\theta_G, \theta_{D2}) \\
\end{split}
\end{equation}
\vspace{-2mm}

\noindent
Here $\lambda_{L1}, \lambda_{MS-SSIM}, \lambda_{SCAL}$ represent weights for balancing the L1, MS-SSIM, and SCAL losses and are set to 1, 1, and 0.01, whereas $\lambda_{P-ADV}, \lambda_{F-ADV}$ represent the weights for balancing the adversarial losses and are set to 0.5. We refer to the generator trained using this process as AFNet.

\vspace{-4mm}
\section{Experimental Evaluations}
\vspace{-2mm}
\subsection{Datasets and Evaluation Metrics}
\vspace{-2mm}
As we analyze the performance of different SoTA image enhancement algorithms along with their application in real perception tasks, we rely upon multiple datasets with different evaluation metrics. Hence we categorize them according to tasks and summarize them as, 

\noindent
\textbf{Image Enhancement} - To examine the performance of SoTA algorithms under diverse illumination conditions, exhaustive experiments are performed using datasets containing both sRGB and RAW images. For sRGB images, we use LOL \cite{wei2018deep} and SICE \cite{Cai2018deep} datasets wherein the LOL dataset contains 1000, 485, 15 paired training, validation, and test images, whereas the SICE dataset contains 400, 130, 58 paired images captured under different illumination conditions ranging from -3ev to +3ev with increments of 1ev. For the RAW dataset, we utilize the SID-Sony \cite{chen2018learning} subset having 2421, 276 training and test samples along with the ELD \cite{wei2020physics} dataset that comprises 384, 96 training and test image pairs captured using different camera sensors. To quantify performance, we use a wide range of metrics covering pixel information (PSNR), structural consistency (SSIM \cite{wang2004image}) and Textural Consistency both with (LPIPS \cite{zhang2018unreasonable}) and without (NIQE \cite{mittal2012making}) reference.

\noindent
\textbf{Perception Algorithms} - To analyze the impact of variable illumination conditions on common perception tasks such as object detection and semantic segmentation, we choose ExDark \cite{Exdark}, JOL \cite{Shyam_2021_ICCV}, COCO \cite{lin2014microsoft} and Cityscapes \cite{Cordts2016Cityscapes} datasets and use mAP and mIOU metrics to quantify performance in dynamic illumination conditions. While Exdark and JOL captures dynamic illumination conditions for object detection, we extend the COCO and Cityscapes datasets to represent night conditions using image translation methods to verify the results across datasets and tasks. Specifically, we improve performance of original Cycle-GAN \cite{zhu2017unpaired} by introducing different techniques which are provided in supplementary. While such techniques could be inversely used to generate well-illuminated images, the computational cost associated with these algorithms to process high-resolution images overshadows their style translation performance.

\vspace{-4mm}
\subsection{Training Mechanism}
\vspace{-2mm}
To enhance sRGB images, we utilized images from the LOL dataset, cropped to 256 x 256 along with augmentation techniques mentioned \cite{shyam2021evaluating} that ensure the presence of different illumination conditions within training samples. We used an ADAM \cite{kingma2014adam} optimizer with an initial learning rate of 1e-4 for both generator and discriminators. We train the complete framework for 1000 epochs while reducing the learning rate by a factor of 0.5 every 200 epochs and use the model weights that result in minimum validation error across the training process. As we study if camera-ISP could be integrated within the enhancement pipeline, we modify all prior sRGB algorithms to accept 4 channel demosaiced input and use bicubic upsampling to match the resolution of generated images with ground truth. Furthermore, we follow the training process mentioned above without making any modifications to the underlying CNN architecture. For our experiments, we use a system equipped with NVidia 3090 GPU running Pytorch 1.7.

\begin{figure}[!t]
\centering
\scriptsize
\renewcommand{\tabcolsep}{3pt} 
\begin{minipage}{0.475\linewidth}\centering
\renewcommand{\tabcolsep}{2pt} 
\renewcommand{\arraystretch}{1} 
\captionof{table}{Quantitative results of ablation studies.} \vspace{2mm}
\begin{adjustbox}{width=\columnwidth}
\begin{tabular}{l c c}
\Xhline{3\arrayrulewidth} \hline \noalign{\vskip 1pt}
Config. & PSNR / SSIM & GMACs\\ 
\Xhline{2\arrayrulewidth} \hline \noalign{\vskip 1pt}
Single Stage                 & 14.25 / 0.57 & 0.46 \\
Two Stage w 1x RDB           & 18.16 / 0.62 & 0.57 \\
w 3x RDB                     & 19.67 / 0.64 & 0.92 \\
w 5x RDB                     & 20.48 / 0.69 & 1.58 \\
w 7x RDB                     & 20.51 / 0.71 & 2.23 \\
Two Stage w 7x RDB + cMSFE-A & 21.45 / 0.78 & 4.38 \\
+ Patch GAN                  & 21.97 / 0.82 & 4.38 \\
+ Fourier GAN (RGB)          & 22.98 / 0.83 & 4.38 \\
AFNet (+ Fourier GAN (Gray)) & 23.01 / 0.84 & 4.38 \\

\Xhline{2\arrayrulewidth} \hline
\end{tabular}
\end{adjustbox}
\label{tab_3}
\end{minipage}
\quad
\begin{minipage}{0.475\linewidth}
\includegraphics[width=\columnwidth, height=5cm]{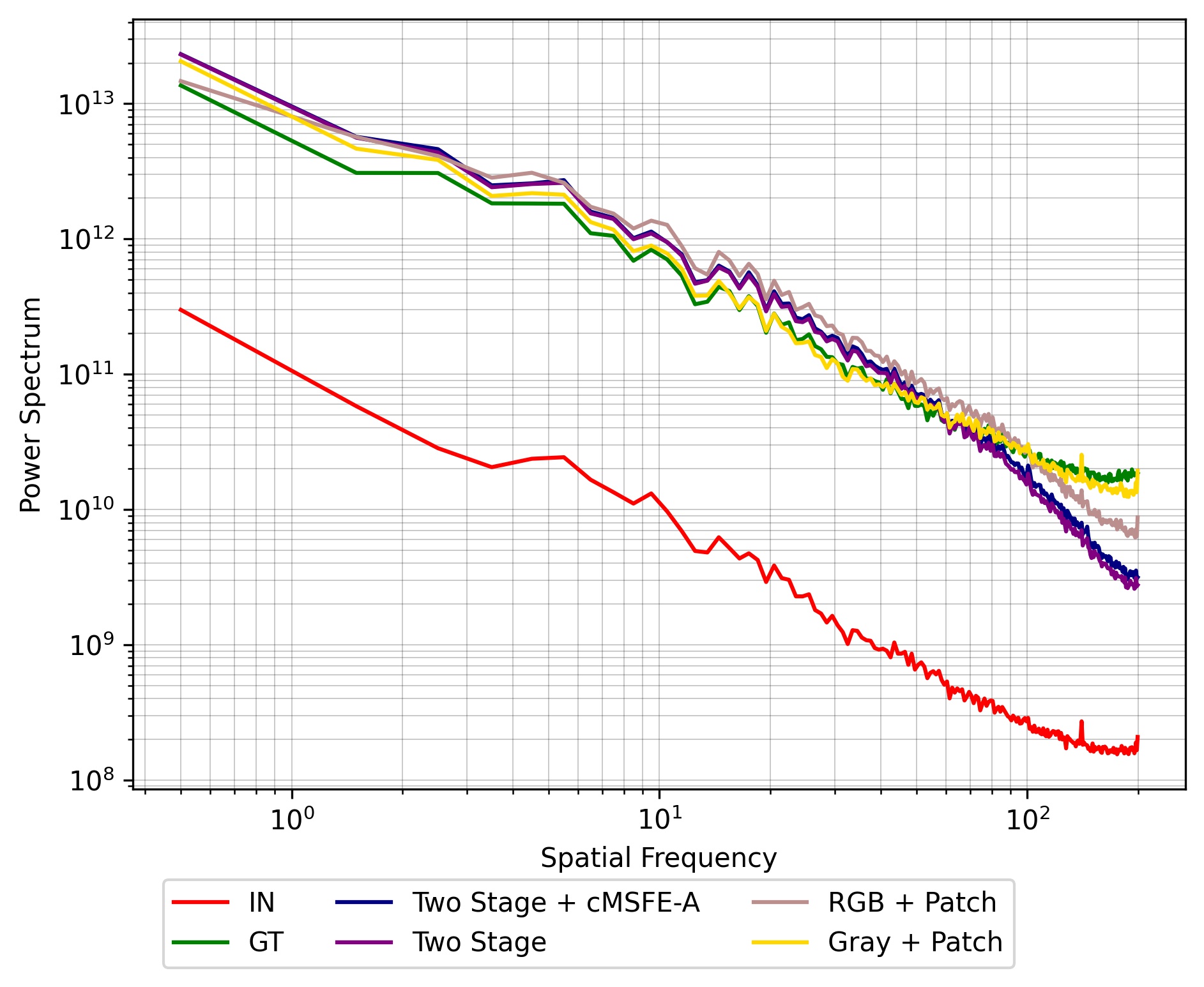} \\
\caption{Power Spectral Density (PSD) of images generated using different configurations of proposed framework.}
\label{fig_5}
\end{minipage}
\vspace{-6mm}
\end{figure}

\vspace{-4mm}
\subsection{Ablation Studies}
\vspace{-2mm}
In this section, we examine the effect of different architectural and optimization modifications on network performance in terms of computational cost as well as the quality of the enhanced image. To carry out our examination, we use the LOL dataset as its small size allows us to explore different network variants while minimizing the training time. We first compare the performance of single-stage and two-stage networks that are trained in an end-to-end manner without using pixel, structural and perceptual losses. From performance results summarized in Tab. \ref{tab_3} and PSD in Fig.,\ref{fig_5} we observe the two-stage network with 5 RDB blocks to result in peak performance in PSNR and SSIM. We further replace the standard convolutional layers with proposed cMSFE-A layers and observe performance to improve significantly both in terms of quantitative metrics and the PSD curves as well with a large textural component matching the ground truth PSD. Subsequently, we examine the effect of training proposed two-stage enhancement algorithm in GAN framework, specifically focusing on the effect of Fourier spectrum-based discriminator on quality of generated images. Quantitative and qualitative results (In Supplementary) demonstrate that using Fourier discriminator improves image generation quality, while the method of extracting Fourier spectrum from a grayscale image or per-channel of RGB image doesn't make a significant difference on performance. From the PSD curves, we can verify that using Fourier adversarial networks indeed improves the performance of enhancement algorithms in the higher frequency spectrum. (We present extended analysis in Supplementary).

\begin{table}
\centering
\scriptsize
\renewcommand{\tabcolsep}{3pt} 
\def\arraystretch{1.2}
\begin{minipage}{0.475\linewidth}\centering
\caption{Performance Evaluation of SoTA on sRGB images from LOL dataset.} \vspace{2mm}
\begin{adjustbox}{width=\columnwidth}
\begin{tabular}{l c c c}
\Xhline{2\arrayrulewidth} \hline \noalign{\vskip 1pt}
Algorithm & PSNR / SSIM ($\uparrow$) & NIQE / LPIPS ($\downarrow$) & GMACs  \\ 
\Xhline{2\arrayrulewidth} \hline \noalign{\vskip 1pt}
Input           &  7.77 / 0.19 & 5.71 / 0.42 & - \\
DALE            & 18.55 / 0.73 & 9.43 / 0.28 & 211.47  \\ 
DLN             & 21.34 / 0.82 & 3.05 / 0.28 & 248.02  \\ 
DSLR            & 18.22 / 0.62 & 3.90 / 0.58 &  18.74  \\ 
EnlightenGAN    & 17.48 / 0.65 & 4.89 / 0.39 &  61.07  \\ 
GLAD            & 19.72 / 0.68 & 6.80 / 0.40 &   0.12  \\ 
KinD            & 17.65 / 0.77 & 3.89 / 0.28 &  14.62  \\ 
MBLLEN          & 17.63 / 0.72 & 3.38 / 0.37 & 104.76  \\ 
RetinexNet      & 16.77 / 0.42 & 9.73 / 0.47 &  68.00  \\ 
URIE            & 20.10 / 0.72 & 4.75 / 0.41 &  14.28  \\ 
Ours            & 23.01 / 0.84 & 3.86 / 0.27 &   4.38  \\
\Xhline{2\arrayrulewidth} \hline \noalign{\vskip 1pt}
\end{tabular}
\end{adjustbox}
\label{tab_1}
\end{minipage}
\quad 
\begin{minipage}{0.475\linewidth}\centering
\caption{Performance Evaluation of SoTA on RAW images from SID-Sony dataset.} \vspace{2mm}
\begin{adjustbox}{width=\columnwidth}
\begin{tabular}{l c c c}
\Xhline{2\arrayrulewidth} \hline \noalign{\vskip 1pt}
Algorithm & PSNR / SSIM ($\uparrow$) & NIQE / LPIPS ($\downarrow$) & GMACs  \\ 
\Xhline{2\arrayrulewidth} \hline \noalign{\vskip 1pt}
Rawpy           & 28.73 / 0.77 & 4.07 / 0.38 & - \\
EnlightenGAN    & 24.27 / 0.64 & 4.68 / 0.53 & 349.20 \\
RAW2RGB-GAN     & 23.55 / 0.78 & 4.00 / 0.71 & 342.19 \\
KinD            & 26.91 / 0.73 & 4.10 / 0.39 & 196.07 \\ 
GLAD            & 27.11 / 0.82 & 3.86 / 0.39 & 132.64 \\
SID             & 28.88 / 0.78 & 4.39 / 0.43 & 562.06 \\
TENet           & 30.17 / 0.83 & 3.18 / 0.31 & 1560.14 \\
PyNet           & 29.01 / 0.79 & 3.79 / 0.34 & 2097.03 \\
PyNet-CA        & 27.24 / 0.74 & 4.02 / 0.41 & 2194.14 \\
AWNet           & 28.09 / 0.76 & 3.98 / 0.39 & 460.29 \\
Ours            & 27.67 / 0.84 & 3.94 / 0.37 & 168.08 \\
\Xhline{2\arrayrulewidth} \hline \noalign{\vskip 1pt}
\end{tabular}
\end{adjustbox}
\label{tab_2}
\end{minipage}
\vspace{-4mm}
\end{table}

\vspace{-3mm}
\subsection{Performance Evaluation with SoTA Algorithms}
\vspace{-1mm}
For comparing enhancement performance on sRGB images, we choose publicly available supervised-learning-based algorithms such as DALE \cite{kwon2020dale}, DLN \cite{DLN2020}, DSLR \cite{lim2020dslr}, EnlightenGAN \cite{jiang2019enlightengan}, GLAD \cite{wang2018gladnet}, MBLLEN \cite{lv2018mbllen}, KinD \cite{zhang2019kindling}, RetinexNet \cite{wei2018deep} and URIE \cite{son2020urie}, whereas for RAW images, we choose Rawpy \footnote{https://letmaik.github.io/rawpy/api/index.html}, RAW2RGB-GAN \cite{zhao2019saliency}, TENet \cite{qian2019trinity}, PyNet \cite{ignatov2020replacing}, ELD \cite{wei2020physics}, AWNet \cite{dai2020awnet}. We summarize the qualitative performance on LOL and SID-Sony datasets along with computational requirement in GMAC (Giga- Multiplication and Accumulation Operations)\footnote{https://github.com/sovrasov/flops-counter.pytorch} \footnote{Assuming 1 GMACs = 0.5 GFLOPs} in Tab. \ref{tab_1} and Tab. \ref{tab_2}, respectively.

From performance metrics, we can conclude that algorithms comprising multiple subnetworks (GLAD, KinD, DSLR) could provide comparable performance with respect to SoTA (URIE, EnlightenGAN) without consuming excessive computations, with the proposed approach providing new SoTA without excessive computations. In addition, we observe that GAN-based approaches such as EnlightenGAN and AFNet result in improved scores on feature-based metrics such as NIQE and LPIPS, thereby demonstrating GAN-based approaches to generate naturalistic images. In order to examine if these algorithms could be reconfigured to accept demosaiced RAW images and generate enhanced sRGB images, we use EnlightenGAN, KinD, GLAD, and AFNet along with bicubic upsampling mechanism and summarize results in Tab. \ref{tab_2}. We observe the performance of these reconfigured algorithms to reach the performance of algorithms that are specifically designed for RAW image enhancement albeit a lower computational requirement.

\vspace{-4mm}
\section{Qualitative Evaluation}
\vspace{-2mm}

\begin{figure}[!ht]
\centering
\scriptsize
\renewcommand{\tabcolsep}{1pt} 
\renewcommand{\arraystretch}{1} 
\begin{adjustbox}{width=\columnwidth}
\begin{tabular}{cccccc}
\includegraphics[width=0.16\columnwidth, height=2.0cm]{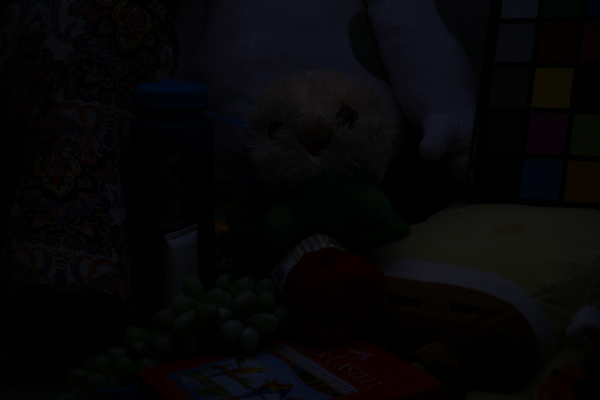} &
\includegraphics[width=0.16\columnwidth, height=2.0cm]{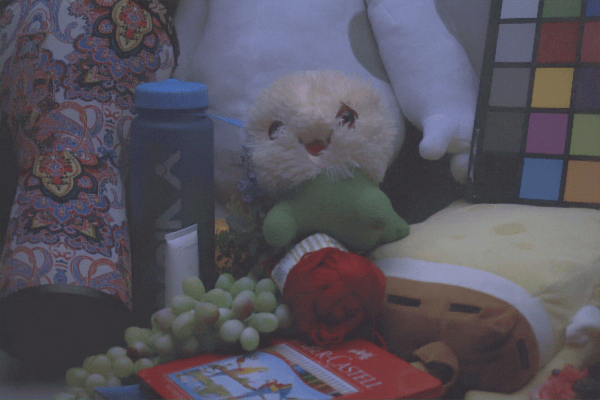} &
\includegraphics[width=0.16\columnwidth, height=2.0cm]{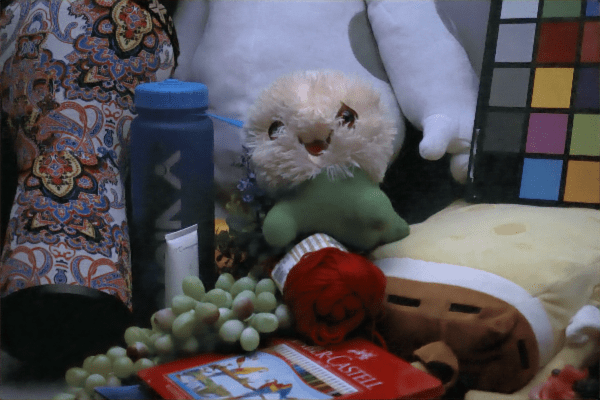} &
\includegraphics[width=0.16\columnwidth, height=2.0cm]{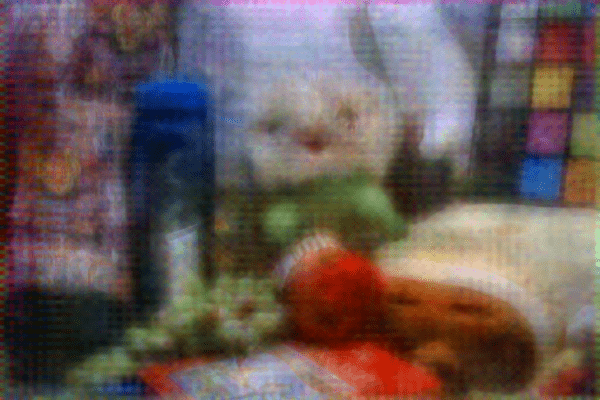} & \includegraphics[width=0.16\columnwidth, height=2.0cm]{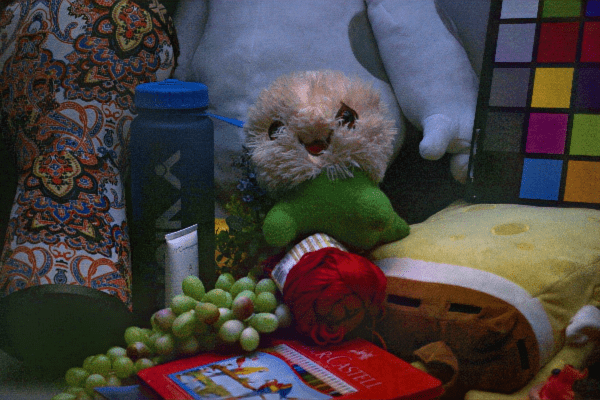} &
\includegraphics[width=0.16\columnwidth, height=2.0cm]{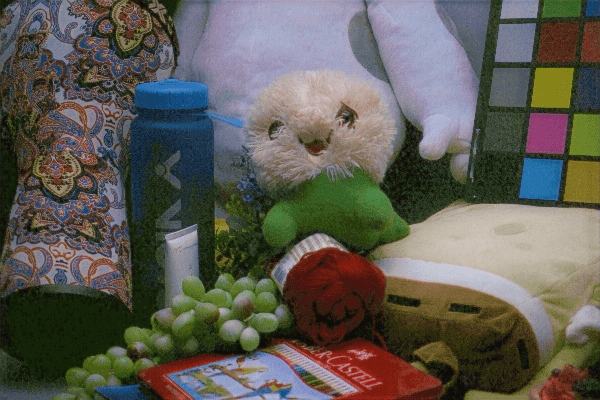} \\
5.01 / 0.46 & 6.67 / 0.35 & 2.73 / 0.29 & 4.80 / 0.64 & 3.42 / 0.29 & 5.18 / 0.38 \\
Input & DALE \cite{kwon2020dale} & DLN \cite{DLN2020} & DSLR \cite{lim2020dslr} & EnlightenGAN \cite{jiang2019enlightengan} & GLAD \cite{wang2018gladnet}  \\

\includegraphics[width=0.16\columnwidth, height=2.0cm]{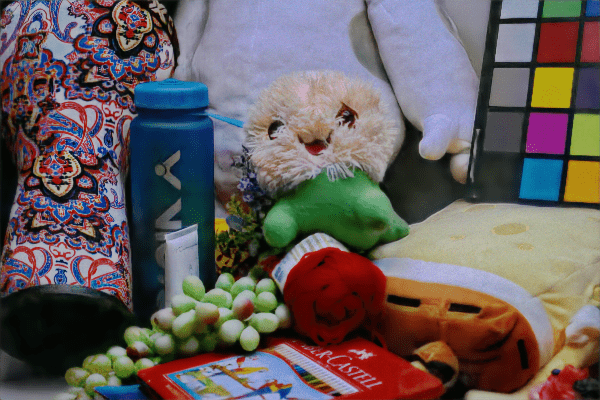} &
\includegraphics[width=0.16\columnwidth, height=2.0cm]{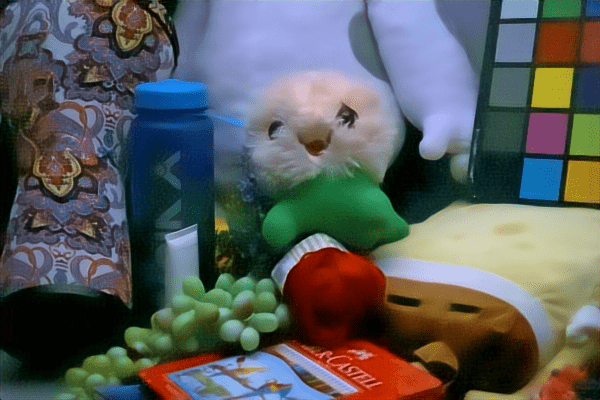} & 
\includegraphics[width=0.16\columnwidth, height=2.0cm]{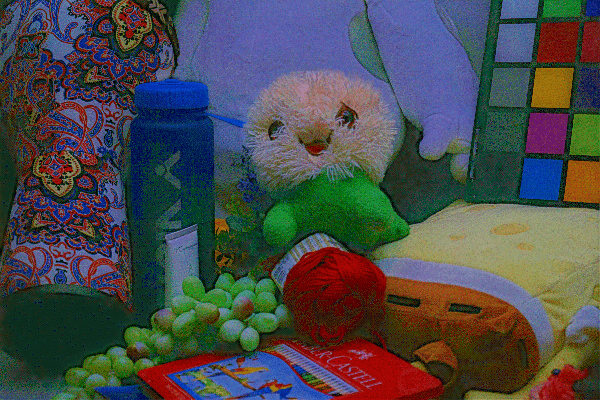} &
\includegraphics[width=0.16\columnwidth, height=2.0cm]{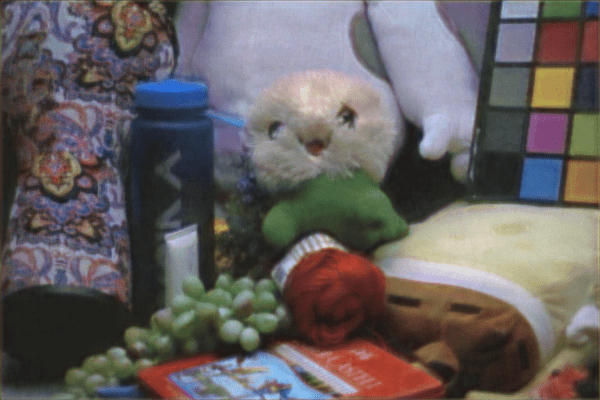} &
\includegraphics[width=0.16\columnwidth, height=2.0cm]{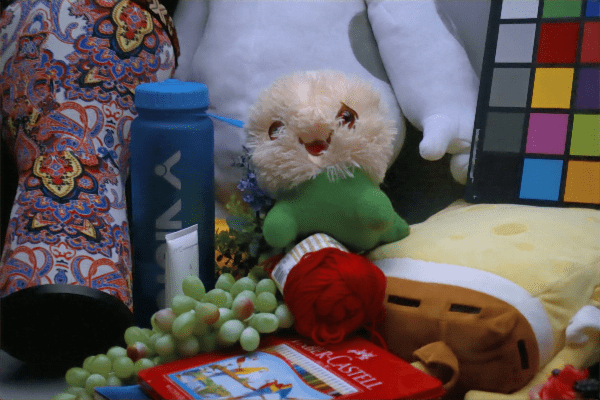} &
\includegraphics[width=0.16\columnwidth, height=2.0cm]{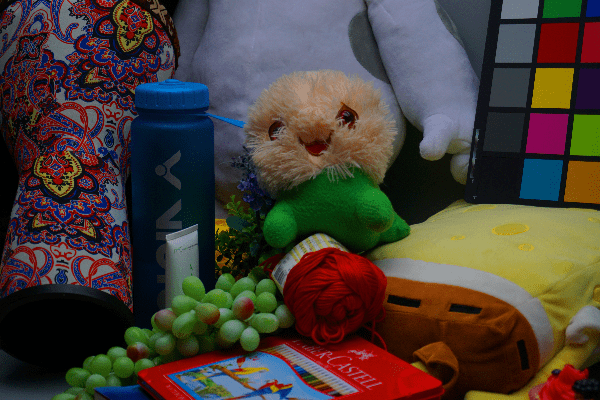} \\
2.94 / 0.27 & 2.84 / 0.39 & 8.49 / 0.38 & 3.62 / 0.43 & 2.91 / 0.18 & 3.87 / - \\ 
KinD \cite{zhang2019kindling} & MBLLEN \cite{lv2018mbllen} & RetinexNet \cite{wei2018deep} & URIE \cite{son2020urie} & AFNet & GT \\ \\
\end{tabular}
\end{adjustbox}
\caption{Performance of SoTA algorithms on image from LOL dataset with NIQE / LPIPS score respectively.}
\label{fig_7}
\vspace{-4mm}
\end{figure}

We present some visual results demonstrating the effectiveness of proposed approach for qualitative evaluation, with additional examples included in supplementary material. Specifically we show performance comparison with SoTA algorithms on LOL dataset in Fig. \ref{fig_7} along with performance of SoTA object detection algorithms on low light and enhanced images from ExDark dataset \cite{Exdark}. From these results we demonstrate both quantitative and qualitative superiority of the proposed mechanism that aids in performance of SoTA Object detection algorithms. 

\begin{figure}[!ht]
\centering
\scriptsize
\renewcommand{\tabcolsep}{1pt} 
\renewcommand{\arraystretch}{1} 
\begin{adjustbox}{width=\columnwidth}
\begin{tabular}{ccc}
\includegraphics[width=0.3\columnwidth, height=1.5cm]{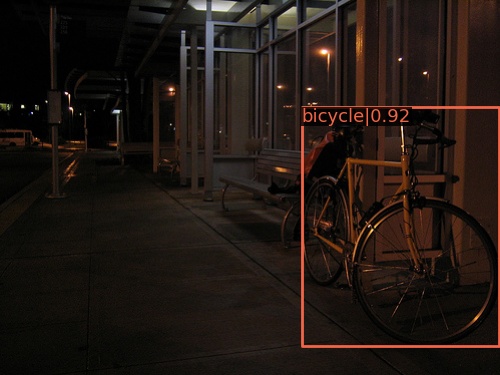} &
\includegraphics[width=0.3\columnwidth, height=1.5cm]{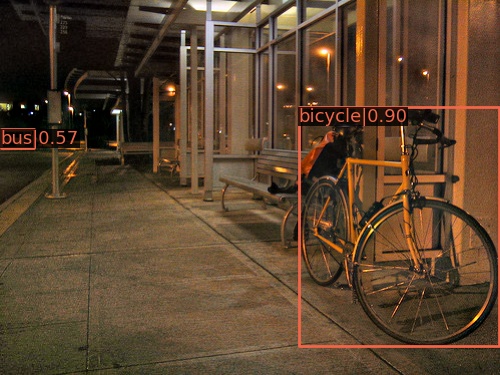} &
\includegraphics[width=0.3\columnwidth, height=1.5cm]{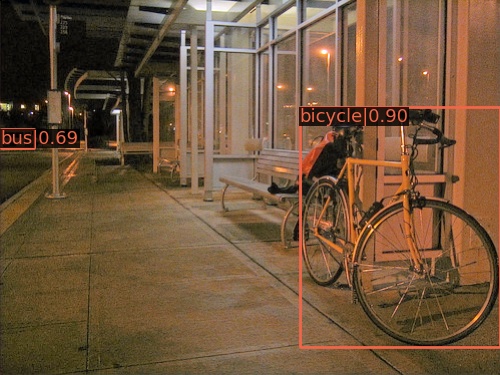} \\
LL Image & EnlightenGAN \cite{jiang2019enlightengan} & DLN \cite{DLN2020} \\
\includegraphics[width=0.3\columnwidth, height=1.5cm]{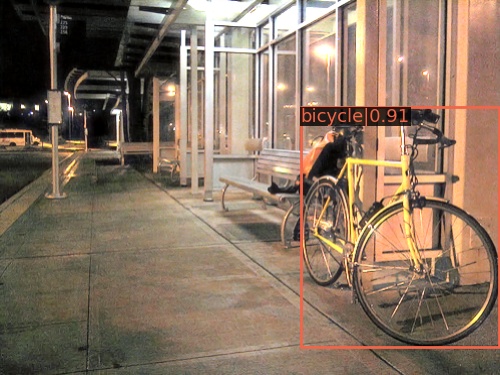} &
\includegraphics[width=0.3\columnwidth, height=1.5cm]{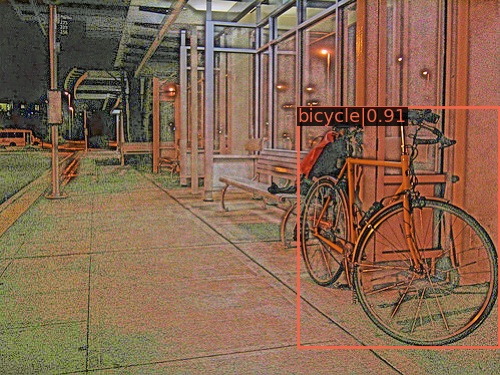} &
\includegraphics[width=0.3\columnwidth, height=1.5cm]{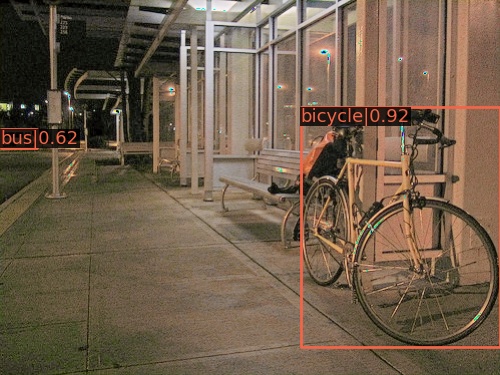} \\ 
GLAD \cite{wang2018gladnet} & RetinexNet \cite{wei2018deep} & AFNet \\ \\
\end{tabular}
\end{adjustbox}
\caption{Qualitative Performance of Deformable DETR \cite{zhu2021deformable} object detector on Low Light and Enhanced Images following different SoTA.}
\label{fig_13}
\vspace{-4mm}
\end{figure}


\vspace{-5mm}
\section{Conclusion}
\vspace{-2mm}
In this paper, we presented the need for balancing illumination present in an image and proposed a fourier adversarial network to ensure presence of structural details within enhanced images. Subsequently we demonstrated the proposed approach to provide SoTA performance while consuming minimum computational resources making it lucrative to be deployed on edge or resource constrained devices. We further demonstrated that Camera-ISP adversely affects the performance of image enhancement algorithms, which can be improved if raw demosaiced images are used as inputs thus the image enhancement algorithm can integrate the functionality of camera-ISP. Finally we demonstrate the varying illumination conditions adversely affect the performance of SoTA object detection and semantic segmentation algorithms which can be improved using image enhancement algorithms. 

\noindent
\textbf{Acknowledgement} This research was supported by KAIST-KU Joint Research Center, KAIST, Korea (N11200035).

\clearpage
\bibliography{bmvc_arxiv}

\begin{thebibliography}{86}
\providecommand{\natexlab}[1]{#1}
\providecommand{\url}[1]{\texttt{#1}}
\expandafter\ifx\csname urlstyle\endcsname\relax
  \providecommand{\doi}[1]{doi: #1}\else
  \providecommand{\doi}{doi: \begingroup \urlstyle{rm}\Url}\fi

\bibitem[Abdelhamed et~al.(2020)Abdelhamed, Afifi, Timofte, and
  Brown]{abdelhamed2020ntire}
Abdelrahman Abdelhamed, Mahmoud Afifi, Radu Timofte, and Michael~S Brown.
\newblock Ntire 2020 challenge on real image denoising: Dataset, methods and
  results.
\newblock In \emph{Proceedings of the IEEE/CVF Conference on Computer Vision
  and Pattern Recognition Workshops}, pages 496--497, 2020.

\bibitem[Bychkovsky et~al.(2011)Bychkovsky, Paris, Chan, and Durand]{fivek}
Vladimir Bychkovsky, Sylvain Paris, Eric Chan, and Fr{\'e}do Durand.
\newblock Learning photographic global tonal adjustment with a database of
  input / output image pairs.
\newblock In \emph{The Twenty-Fourth IEEE Conference on Computer Vision and
  Pattern Recognition}, 2011.

\bibitem[Cai et~al.(2018)Cai, Gu, and Zhang]{Cai2018deep}
Jianrui Cai, Shuhang Gu, and Lei Zhang.
\newblock Learning a deep single image contrast enhancer from multi-exposure
  images.
\newblock \emph{IEEE Transactions on Image Processing}, 27\penalty0
  (4):\penalty0 2049--2062, 2018.

\bibitem[Chen et~al.(2018{\natexlab{a}})Chen, Chen, Xu, and
  Koltun]{chen2018learning}
Chen Chen, Qifeng Chen, Jia Xu, and Vladlen Koltun.
\newblock Learning to see in the dark.
\newblock In \emph{Proceedings of the IEEE Conference on Computer Vision and
  Pattern Recognition}, pages 3291--3300, 2018{\natexlab{a}}.

\bibitem[Chen et~al.(2019{\natexlab{a}})Chen, Pang, Wang, Xiong, Li, Sun, Feng,
  Liu, Shi, Ouyang, Loy, and Lin]{chen2019hybrid}
Kai Chen, Jiangmiao Pang, Jiaqi Wang, Yu~Xiong, Xiaoxiao Li, Shuyang Sun,
  Wansen Feng, Ziwei Liu, Jianping Shi, Wanli Ouyang, Chen~Change Loy, and
  Dahua Lin.
\newblock Hybrid task cascade for instance segmentation.
\newblock In \emph{IEEE Conference on Computer Vision and Pattern Recognition},
  2019{\natexlab{a}}.

\bibitem[Chen et~al.(2019{\natexlab{b}})Chen, Wang, Pang, Cao, Xiong, Li, Sun,
  Feng, Liu, Xu, Zhang, Cheng, Zhu, Cheng, Zhao, Li, Lu, Zhu, Wu, Dai, Wang,
  Shi, Ouyang, Loy, and Lin]{mmdetection}
Kai Chen, Jiaqi Wang, Jiangmiao Pang, Yuhang Cao, Yu~Xiong, Xiaoxiao Li,
  Shuyang Sun, Wansen Feng, Ziwei Liu, Jiarui Xu, Zheng Zhang, Dazhi Cheng,
  Chenchen Zhu, Tianheng Cheng, Qijie Zhao, Buyu Li, Xin Lu, Rui Zhu, Yue Wu,
  Jifeng Dai, Jingdong Wang, Jianping Shi, Wanli Ouyang, Chen~Change Loy, and
  Dahua Lin.
\newblock {MMDetection}: Open mmlab detection toolbox and benchmark.
\newblock \emph{arXiv preprint arXiv:1906.07155}, 2019{\natexlab{b}}.

\bibitem[Chen et~al.(2018{\natexlab{b}})Chen, Zhu, Papandreou, Schroff, and
  Adam]{deeplabv3plus2018}
Liang-Chieh Chen, Yukun Zhu, George Papandreou, Florian Schroff, and Hartwig
  Adam.
\newblock Encoder-decoder with atrous separable convolution for semantic image
  segmentation.
\newblock In \emph{ECCV}, 2018{\natexlab{b}}.

\bibitem[Chen et~al.(2021)Chen, Wang, Yang, Zhang, Cheng, and Sun]{chen2021you}
Qiang Chen, Yingming Wang, Tong Yang, Xiangyu Zhang, Jian Cheng, and Jian Sun.
\newblock You only look one-level feature.
\newblock In \emph{IEEE Conference on Computer Vision and Pattern Recognition},
  2021.

\bibitem[Contributors(2020)]{mmseg2020}
MMSegmentation Contributors.
\newblock {MMSegmentation}: Openmmlab semantic segmentation toolbox and
  benchmark.
\newblock \url{https://github.com/open-mmlab/mmsegmentation}, 2020.

\bibitem[Cordts et~al.(2016)Cordts, Omran, Ramos, Rehfeld, Enzweiler, Benenson,
  Franke, Roth, and Schiele]{Cordts2016Cityscapes}
Marius Cordts, Mohamed Omran, Sebastian Ramos, Timo Rehfeld, Markus Enzweiler,
  Rodrigo Benenson, Uwe Franke, Stefan Roth, and Bernt Schiele.
\newblock The cityscapes dataset for semantic urban scene understanding.
\newblock In \emph{Proc. of the IEEE Conference on Computer Vision and Pattern
  Recognition (CVPR)}, 2016.

\bibitem[Dabov et~al.(2007)Dabov, Foi, Katkovnik, and
  Egiazarian]{dabov2007image}
Kostadin Dabov, Alessandro Foi, Vladimir Katkovnik, and Karen Egiazarian.
\newblock Image denoising by sparse 3-d transform-domain collaborative
  filtering.
\newblock \emph{IEEE Transactions on image processing}, 16\penalty0
  (8):\penalty0 2080--2095, 2007.

\bibitem[Dai et~al.(2020)Dai, Liu, Li, and Chen]{dai2020awnet}
Linhui Dai, Xiaohong Liu, Chengqi Li, and Jun Chen.
\newblock Awnet: Attentive wavelet network for image isp.
\newblock \emph{arXiv preprint arXiv:2008.09228}, 2020.

\bibitem[Guan et~al.(2019)Guan, Liu, Moran, Song, and Slabaugh]{guan2019node}
Hao Guan, Liu Liu, Sean Moran, Fenglong Song, and Gregory Slabaugh.
\newblock Node: Extreme low light raw image denoising using a noise
  decomposition network.
\newblock \emph{arXiv preprint arXiv:1909.05249}, 2019.

\bibitem[He et~al.(2019)He, Deng, Zhou, Wang, and Qiao]{He_2019_CVPR}
Junjun He, Zhongying Deng, Lei Zhou, Yali Wang, and Yu~Qiao.
\newblock Adaptive pyramid context network for semantic segmentation.
\newblock In \emph{Proceedings of the IEEE/CVF Conference on Computer Vision
  and Pattern Recognition (CVPR)}, June 2019.

\bibitem[Hu et~al.(2018)Hu, Shen, and Sun]{hu2018squeeze}
Jie Hu, Li~Shen, and Gang Sun.
\newblock Squeeze-and-excitation networks.
\newblock In \emph{Proceedings of the IEEE conference on computer vision and
  pattern recognition}, pages 7132--7141, 2018.

\bibitem[Huang et~al.(2019)Huang, Wang, Huang, Huang, Wei, and
  Liu]{huang2018ccnet}
Zilong Huang, Xinggang Wang, Lichao Huang, Chang Huang, Yunchao Wei, and Wenyu
  Liu.
\newblock Ccnet: Criss-cross attention for semantic segmentation.
\newblock 2019.

\bibitem[Ignatov et~al.(2020)Ignatov, Van~Gool, and
  Timofte]{ignatov2020replacing}
Andrey Ignatov, Luc Van~Gool, and Radu Timofte.
\newblock Replacing mobile camera isp with a single deep learning model.
\newblock \emph{arXiv preprint arXiv:2002.05509}, 2020.

\bibitem[Isola et~al.(2017{\natexlab{a}})Isola, Zhu, Zhou, and
  Efros]{isola2017image}
Phillip Isola, Jun-Yan Zhu, Tinghui Zhou, and Alexei~A Efros.
\newblock Image-to-image translation with conditional adversarial networks.
\newblock In \emph{Proceedings of the IEEE conference on computer vision and
  pattern recognition}, pages 1125--1134, 2017{\natexlab{a}}.

\bibitem[Isola et~al.(2017{\natexlab{b}})Isola, Zhu, Zhou, and
  Efros]{pix2pix2017}
Phillip Isola, Jun-Yan Zhu, Tinghui Zhou, and Alexei~A Efros.
\newblock Image-to-image translation with conditional adversarial networks.
\newblock \emph{CVPR}, 2017{\natexlab{b}}.

\bibitem[Jiang et~al.(2019)Jiang, Gong, Liu, Cheng, Fang, Shen, Yang, Zhou, and
  Wang]{jiang2019enlightengan}
Yifan Jiang, Xinyu Gong, Ding Liu, Yu~Cheng, Chen Fang, Xiaohui Shen, Jianchao
  Yang, Pan Zhou, and Zhangyang Wang.
\newblock Enlightengan: Deep light enhancement without paired supervision.
\newblock \emph{arXiv preprint arXiv:1906.06972}, 2019.

\bibitem[Karras et~al.(2020)Karras, Aittala, Hellsten, Laine, Lehtinen, and
  Aila]{karras2020training}
Tero Karras, Miika Aittala, Janne Hellsten, Samuli Laine, Jaakko Lehtinen, and
  Timo Aila.
\newblock Training generative adversarial networks with limited data.
\newblock \emph{arXiv preprint arXiv:2006.06676}, 2020.

\bibitem[Khosla et~al.(2020)Khosla, Teterwak, Wang, Sarna, Tian, Isola,
  Maschinot, Liu, and Krishnan]{khosla2020supervised}
Prannay Khosla, Piotr Teterwak, Chen Wang, Aaron Sarna, Yonglong Tian, Phillip
  Isola, Aaron Maschinot, Ce~Liu, and Dilip Krishnan.
\newblock Supervised contrastive learning.
\newblock \emph{arXiv preprint arXiv:2004.11362}, 2020.

\bibitem[Kim and Lee(2020)]{paa-eccv2020}
Kang Kim and Hee~Seok Lee.
\newblock Probabilistic anchor assignment with iou prediction for object
  detection.
\newblock In \emph{ECCV}, 2020.

\bibitem[Kingma and Ba(2014)]{kingma2014adam}
Diederik~P Kingma and Jimmy Ba.
\newblock Adam: A method for stochastic optimization.
\newblock \emph{arXiv preprint arXiv:1412.6980}, 2014.

\bibitem[Kirillov et~al.(2019)Kirillov, Girshick, He, and
  Dollar]{Kirillov_2019}
Alexander Kirillov, Ross Girshick, Kaiming He, and Piotr Dollar.
\newblock Panoptic feature pyramid networks.
\newblock \emph{2019 IEEE/CVF Conference on Computer Vision and Pattern
  Recognition (CVPR)}, Jun 2019.
\newblock \doi{10.1109/cvpr.2019.00656}.
\newblock URL \url{http://dx.doi.org/10.1109/CVPR.2019.00656}.

\bibitem[Kirillov et~al.(2020)Kirillov, Wu, He, and
  Girshick]{kirillov2020pointrend}
Alexander Kirillov, Yuxin Wu, Kaiming He, and Ross Girshick.
\newblock Pointrend: Image segmentation as rendering.
\newblock In \emph{Proceedings of the IEEE/CVF conference on computer vision
  and pattern recognition}, pages 9799--9808, 2020.

\bibitem[Kokkinos and Lefkimmiatis(2019)]{kokkinos2019iterative}
Filippos Kokkinos and Stamatios Lefkimmiatis.
\newblock Iterative joint image demosaicking and denoising using a residual
  denoising network.
\newblock \emph{IEEE Transactions on Image Processing}, 28\penalty0
  (8):\penalty0 4177--4188, 2019.

\bibitem[Kwon et~al.(2020)Kwon, Kim, and Kwon]{kwon2020dale}
Dokyeong Kwon, Guisik Kim, and Junseok Kwon.
\newblock Dale: Dark region-aware low-light image enhancement.
\newblock \emph{arXiv preprint arXiv:2008.12493}, 2020.

\bibitem[Lee et~al.(2016)Lee, Hirakawa, and Nguyen]{lee2016joint}
Yeejin Lee, Keigo Hirakawa, and Truong~Q Nguyen.
\newblock Joint defogging and demosaicking.
\newblock \emph{IEEE Transactions on Image Processing}, 26\penalty0
  (6):\penalty0 3051--3063, 2016.

\bibitem[Li et~al.(2020)Li, Wang, Wu, Chen, Hu, Li, Tang, and
  Yang]{li2020generalized}
Xiang Li, Wenhai Wang, Lijun Wu, Shuo Chen, Xiaolin Hu, Jun Li, Jinhui Tang,
  and Jian Yang.
\newblock Generalized focal loss: Learning qualified and distributed bounding
  boxes for dense object detection.
\newblock \emph{arXiv preprint arXiv:2006.04388}, 2020.

\bibitem[Liang et~al.(2019)Liang, Cai, Cao, and Zhang]{liang2019cameranet}
Zhetong Liang, Jianrui Cai, Zisheng Cao, and Lei Zhang.
\newblock Cameranet: A two-stage framework for effective camera isp learning.
\newblock \emph{arXiv preprint arXiv:1908.01481}, 2019.

\bibitem[Lim and Kim(2020)]{lim2020dslr}
Seokjae Lim and Wonjun Kim.
\newblock Dslr: Deep stacked laplacian restorer for low-light image
  enhancement.
\newblock \emph{IEEE Transactions on Multimedia}, 2020.

\bibitem[Lin et~al.(2013)Lin, Chen, and Yan]{lin2013network}
Min Lin, Qiang Chen, and Shuicheng Yan.
\newblock Network in network.
\newblock \emph{arXiv preprint arXiv:1312.4400}, 2013.

\bibitem[Lin et~al.(2014)Lin, Maire, Belongie, Hays, Perona, Ramanan,
  Doll{\'a}r, and Zitnick]{lin2014microsoft}
Tsung-Yi Lin, Michael Maire, Serge Belongie, James Hays, Pietro Perona, Deva
  Ramanan, Piotr Doll{\'a}r, and C~Lawrence Zitnick.
\newblock Microsoft coco: Common objects in context.
\newblock In \emph{European conference on computer vision}, pages 740--755.
  Springer, 2014.

\bibitem[Lin et~al.(2017)Lin, Goyal, Girshick, He, and
  Doll{\'a}r]{lin2017focal}
Tsung-Yi Lin, Priya Goyal, Ross Girshick, Kaiming He, and Piotr Doll{\'a}r.
\newblock Focal loss for dense object detection.
\newblock In \emph{Proceedings of the IEEE international conference on computer
  vision}, 2017.

\bibitem[Liu et~al.(2018)Liu, Zhang, Zhang, Lin, and
  Zuo]{Liu_2018_CVPR_Workshops}
Pengju Liu, Hongzhi Zhang, Kai Zhang, Liang Lin, and Wangmeng Zuo.
\newblock Multi-level wavelet-cnn for image restoration.
\newblock In \emph{The IEEE Conference on Computer Vision and Pattern
  Recognition (CVPR) Workshops}, June 2018.

\bibitem[Liu et~al.(2020)Liu, Chen, Yang, Wu, and Li]{liu2020lane}
Tong Liu, Zhaowei Chen, Yi~Yang, Zehao Wu, and Haowei Li.
\newblock Lane detection in low-light conditions using an efficient data
  enhancement: Light conditions style transfer.
\newblock \emph{arXiv preprint arXiv:2002.01177}, 2020.

\bibitem[Liu et~al.(2016)Liu, Anguelov, Erhan, Szegedy, Reed, Fu, and
  Berg]{Liu_2016}
Wei Liu, Dragomir Anguelov, Dumitru Erhan, Christian Szegedy, Scott Reed,
  Cheng-Yang Fu, and Alexander~C. Berg.
\newblock Ssd: Single shot multibox detector.
\newblock \emph{ECCV}, 2016.

\bibitem[Loh and Chan(2019)]{Exdark}
Yuen~Peng Loh and Chee~Seng Chan.
\newblock Getting to know low-light images with the exclusively dark dataset.
\newblock \emph{Computer Vision and Image Understanding}, 178:\penalty0 30--42,
  2019.
\newblock \doi{https://doi.org/10.1016/j.cviu.2018.10.010}.

\bibitem[Lore et~al.(2017)Lore, Akintayo, and Sarkar]{lore2017llnet}
Kin~Gwn Lore, Adedotun Akintayo, and Soumik Sarkar.
\newblock Llnet: A deep autoencoder approach to natural low-light image
  enhancement.
\newblock \emph{Pattern Recognition}, 61:\penalty0 650--662, 2017.

\bibitem[Lv et~al.(2018)Lv, Lu, Wu, and Lim]{lv2018mbllen}
Feifan Lv, Feng Lu, Jianhua Wu, and Chongsoon Lim.
\newblock Mbllen: Low-light image/video enhancement using cnns.
\newblock In \emph{BMVC}, page 220, 2018.

\bibitem[Mao et~al.(2017)Mao, Li, Xie, Lau, Wang, and
  Paul~Smolley]{mao2017least}
Xudong Mao, Qing Li, Haoran Xie, Raymond~YK Lau, Zhen Wang, and Stephen
  Paul~Smolley.
\newblock Least squares generative adversarial networks.
\newblock In \emph{Proceedings of the IEEE international conference on computer
  vision}, pages 2794--2802, 2017.

\bibitem[Michaelis et~al.(2019)Michaelis, Mitzkus, Geirhos, Rusak, Bringmann,
  Ecker, Bethge, and Brendel]{michaelis2019benchmarking}
Claudio Michaelis, Benjamin Mitzkus, Robert Geirhos, Evgenia Rusak, Oliver
  Bringmann, Alexander~S Ecker, Matthias Bethge, and Wieland Brendel.
\newblock Benchmarking robustness in object detection: Autonomous driving when
  winter is coming.
\newblock \emph{arXiv preprint arXiv:1907.07484}, 2019.

\bibitem[Mittal et~al.(2012)Mittal, Soundararajan, and Bovik]{mittal2012making}
Anish Mittal, Rajiv Soundararajan, and Alan~C Bovik.
\newblock Making a “completely blind” image quality analyzer.
\newblock \emph{IEEE Signal processing letters}, 20\penalty0 (3):\penalty0
  209--212, 2012.

\bibitem[Pizzati et~al.(2021)Pizzati, Cerri, and
  de~Charette]{pizzati2021comogan}
Fabio Pizzati, Pietro Cerri, and Raoul de~Charette.
\newblock {CoMoGAN}: continuous model-guided image-to-image translation.
\newblock In \emph{CVPR}, 2021.

\bibitem[Qian et~al.(2019)Qian, Gu, Ren, Dong, Zhao, and Lin]{qian2019trinity}
Guocheng Qian, Jinjin Gu, Jimmy~S Ren, Chao Dong, Furong Zhao, and Juan Lin.
\newblock Trinity of pixel enhancement: a joint solution for demosaicking,
  denoising and super-resolution.
\newblock \emph{arXiv preprint arXiv:1905.02538}, 2019.

\bibitem[Redmon and Farhadi(2018)]{redmon2018yolov3}
Joseph Redmon and Ali Farhadi.
\newblock Yolov3: An incremental improvement, 2018.

\bibitem[Remez et~al.(2017)Remez, Litany, Giryes, and Bronstein]{remez2017deep}
Tal Remez, Or~Litany, Raja Giryes, and Alex~M Bronstein.
\newblock Deep class-aware image denoising.
\newblock In \emph{2017 international conference on sampling theory and
  applications (SampTA)}, pages 138--142. IEEE, 2017.

\bibitem[Ren et~al.(2017)Ren, He, Girshick, and Sun]{Ren_2017}
Shaoqing Ren, Kaiming He, Ross Girshick, and Jian Sun.
\newblock Faster r-cnn: Towards real-time object detection with region proposal
  networks.
\newblock \emph{IEEE Transactions on Pattern Analysis and Machine
  Intelligence}, Jun 2017.

\bibitem[Ronneberger et~al.(2015)Ronneberger, Fischer, and
  Brox]{ronneberger2015u}
Olaf Ronneberger, Philipp Fischer, and Thomas Brox.
\newblock U-net: Convolutional networks for biomedical image segmentation.
\newblock In \emph{International Conference on Medical image computing and
  computer-assisted intervention}, pages 234--241. Springer, 2015.

\bibitem[Schwartz et~al.(2018)Schwartz, Giryes, and
  Bronstein]{schwartz2018deepisp}
Eli Schwartz, Raja Giryes, and Alex~M Bronstein.
\newblock Deepisp: Toward learning an end-to-end image processing pipeline.
\newblock \emph{IEEE Transactions on Image Processing}, 28\penalty0
  (2):\penalty0 912--923, 2018.

\bibitem[Shyam et~al.(2020{\natexlab{a}})Shyam, Bangunharcana, and
  Kim]{shyam2020retaining}
Pranjay Shyam, Antyanta Bangunharcana, and Kyung-Soo Kim.
\newblock Retaining image feature matching performance under low light
  conditions, 2020{\natexlab{a}}.

\bibitem[Shyam et~al.(2020{\natexlab{b}})Shyam, Yoon, and Kim]{9197076}
Pranjay Shyam, Kuk-Jin Yoon, and Kyung-Soo Kim.
\newblock Dynamic anchor selection for improving object localization.
\newblock In \emph{2020 IEEE International Conference on Robotics and
  Automation (ICRA)}, pages 9477--9483, 2020{\natexlab{b}}.
\newblock \doi{10.1109/ICRA40945.2020.9197076}.

\bibitem[Shyam et~al.(2020{\natexlab{c}})Shyam, Yoon, and
  Kim]{shyam2020towards}
Pranjay Shyam, Kuk-Jin Yoon, and Kyung-Soo Kim.
\newblock Towards domain invariant single image dehazing, 2020{\natexlab{c}}.

\bibitem[Shyam et~al.(2021{\natexlab{a}})Shyam, Sengar, Yoon, and
  Kim]{shyam2021evaluating}
Pranjay Shyam, Sandeep~Singh Sengar, Kuk-Jin Yoon, and Kyung-Soo Kim.
\newblock Evaluating copy-blend augmentation for low level vision tasks.
\newblock \emph{arXiv preprint arXiv:2103.05889}, 2021{\natexlab{a}}.

\bibitem[Shyam et~al.(2021{\natexlab{b}})Shyam, Yoon, and Kim]{Shyam_2021_ICCV}
Pranjay Shyam, Kuk-Jin Yoon, and Kyung-Soo Kim.
\newblock Weakly supervised approach for joint object and lane marking
  detection.
\newblock In \emph{Proceedings of the IEEE/CVF International Conference on
  Computer Vision (ICCV) Workshops}, pages 2885--2895, October
  2021{\natexlab{b}}.

\bibitem[Son et~al.(2020)Son, Kang, Kim, Cho, and Kwak]{son2020urie}
Taeyoung Son, Juwon Kang, Namyup Kim, Sunghyun Cho, and Suha Kwak.
\newblock Urie: Universal image enhancement for visual recognition in the wild.
\newblock In \emph{ECCV}, 2020.

\bibitem[Sun et~al.(2019)Sun, Wang, Yang, and Xiang]{sun2019see}
Lei Sun, Kaiwei Wang, Kailun Yang, and Kaite Xiang.
\newblock See clearer at night: towards robust nighttime semantic segmentation
  through day-night image conversion.
\newblock In \emph{Artificial Intelligence and Machine Learning in Defense
  Applications}, volume 11169, page 111690A. International Society for Optics
  and Photonics, 2019.

\bibitem[Szegedy et~al.(2015)Szegedy, Liu, Jia, Sermanet, Reed, Anguelov,
  Erhan, Vanhoucke, and Rabinovich]{szegedy2015going}
Christian Szegedy, Wei Liu, Yangqing Jia, Pierre Sermanet, Scott Reed, Dragomir
  Anguelov, Dumitru Erhan, Vincent Vanhoucke, and Andrew Rabinovich.
\newblock Going deeper with convolutions.
\newblock In \emph{Proceedings of the IEEE conference on computer vision and
  pattern recognition}, pages 1--9, 2015.

\bibitem[Vu et~al.(2021)Vu, Haeyong, and Yoo]{vu2019cascade}
Thang Vu, Kang Haeyong, and Chang~D Yoo.
\newblock Scnet: Training inference sample consistency for instance
  segmentation.
\newblock In \emph{AAAI}, 2021.

\bibitem[Wang et~al.(2020)Wang, Liu, Siu, and Lun]{DLN2020}
Li-Wen Wang, Zhi-Song Liu, Wan-Chi Siu, and Daniel~P.K. Lun.
\newblock Lightening network for low-light image enhancement.
\newblock \emph{IEEE Transactions on Image Processing}, 2020.
\newblock \doi{10.1109/TIP.2020.3008396}.

\bibitem[Wang et~al.(2018{\natexlab{a}})Wang, Liu, Zhu, Tao, Kautz, and
  Catanzaro]{wang2018pix2pixHD}
Ting-Chun Wang, Ming-Yu Liu, Jun-Yan Zhu, Andrew Tao, Jan Kautz, and Bryan
  Catanzaro.
\newblock High-resolution image synthesis and semantic manipulation with
  conditional gans.
\newblock In \emph{Proceedings of the IEEE Conference on Computer Vision and
  Pattern Recognition}, 2018{\natexlab{a}}.

\bibitem[Wang et~al.(2018{\natexlab{b}})Wang, Wei, Yang, and
  Liu]{wang2018gladnet}
Wenjing Wang, Chen Wei, Wenhan Yang, and Jiaying Liu.
\newblock Gladnet: Low-light enhancement network with global awareness.
\newblock In \emph{Automatic Face \& Gesture Recognition (FG 2018), 2018 13th
  IEEE International Conference}, pages 751--755. IEEE, 2018{\natexlab{b}}.

\bibitem[Wang et~al.(2018{\natexlab{c}})Wang, Girshick, Gupta, and
  He]{wang2018non}
Xiaolong Wang, Ross Girshick, Abhinav Gupta, and Kaiming He.
\newblock Non-local neural networks.
\newblock In \emph{Proceedings of the IEEE conference on computer vision and
  pattern recognition}, pages 7794--7803, 2018{\natexlab{c}}.

\bibitem[{Wang} et~al.(2003){Wang}, {Simoncelli}, and {Bovik}]{1292216}
Z.~{Wang}, E.~P. {Simoncelli}, and A.~C. {Bovik}.
\newblock Multiscale structural similarity for image quality assessment.
\newblock In \emph{The Thrity-Seventh Asilomar Conference on Signals, Systems
  Computers, 2003}, volume~2, pages 1398--1402 Vol.2, 2003.
\newblock \doi{10.1109/ACSSC.2003.1292216}.

\bibitem[Wang et~al.(2004)Wang, Bovik, Sheikh, and Simoncelli]{wang2004image}
Zhou Wang, Alan~C Bovik, Hamid~R Sheikh, and Eero~P Simoncelli.
\newblock Image quality assessment: from error visibility to structural
  similarity.
\newblock \emph{IEEE transactions on image processing}, 13\penalty0
  (4):\penalty0 600--612, 2004.

\bibitem[Wei et~al.(2018)Wei, Wang, Yang, and Liu]{wei2018deep}
Chen Wei, Wenjing Wang, Wenhan Yang, and Jiaying Liu.
\newblock Deep retinex decomposition for low-light enhancement.
\newblock \emph{arXiv preprint arXiv:1808.04560}, 2018.

\bibitem[Wei et~al.(2020)Wei, Fu, Yang, and Huang]{wei2020physics}
Kaixuan Wei, Ying Fu, Jiaolong Yang, and Hua Huang.
\newblock A physics-based noise formation model for extreme low-light raw
  denoising.
\newblock In \emph{IEEE Conference on Computer Vision and Pattern Recognition},
  2020.

\bibitem[Wu et~al.(2020)Wu, Tang, Zhang, Cao, and Zhang]{wu2020cgnet}
Tianyi Wu, Sheng Tang, Rui Zhang, Juan Cao, and Yongdong Zhang.
\newblock Cgnet: A light-weight context guided network for semantic
  segmentation.
\newblock \emph{IEEE Transactions on Image Processing}, 30:\penalty0
  1169--1179, 2020.

\bibitem[Xu et~al.(2020)Xu, Yang, Yin, and Lau]{Xu_2020_CVPR}
Ke~Xu, Xin Yang, Baocai Yin, and Rynson~W.H. Lau.
\newblock Learning to restore low-light images via
  decomposition-and-enhancement.
\newblock In \emph{Proceedings of the IEEE/CVF Conference on Computer Vision
  and Pattern Recognition (CVPR)}, June 2020.

\bibitem[Yin et~al.(2020)Yin, Yao, Cao, Li, Zhang, Lin, and
  Hu]{yin2020disentangled}
Minghao Yin, Zhuliang Yao, Yue Cao, Xiu Li, Zheng Zhang, Stephen Lin, and Han
  Hu.
\newblock Disentangled non-local neural networks, 2020.

\bibitem[Yu et~al.(2020)Yu, Chen, Wang, Xian, Chen, Liu, Madhavan, and
  Darrell]{yu2020bdd100k}
Fisher Yu, Haofeng Chen, Xin Wang, Wenqi Xian, Yingying Chen, Fangchen Liu,
  Vashisht Madhavan, and Trevor Darrell.
\newblock Bdd100k: A diverse driving dataset for heterogeneous multitask
  learning.
\newblock In \emph{Proceedings of the IEEE/CVF conference on computer vision
  and pattern recognition}, pages 2636--2645, 2020.

\bibitem[Yuan et~al.(2020)Yuan, Chen, and Wang]{YuanCW20}
Yuhui Yuan, Xilin Chen, and Jingdong Wang.
\newblock Object-contextual representations for semantic segmentation.
\newblock 2020.

\bibitem[Zamir et~al.(2020)Zamir, Arora, Khan, Hayat, Khan, Yang, and
  Shao]{Zamir2020MIRNet}
Syed~Waqas Zamir, Aditya Arora, Salman Khan, Munawar Hayat, Fahad~Shahbaz Khan,
  Ming-Hsuan Yang, and Ling Shao.
\newblock Learning enriched features for real image restoration and
  enhancement.
\newblock In \emph{ECCV}, 2020.

\bibitem[Zhang et~al.(2020{\natexlab{a}})Zhang, Yan, Zhu, Li, Sun, and
  Zhang]{zhang2020attention}
Cheng Zhang, Qingsen Yan, Yu~Zhu, Xianjun Li, Jinqiu Sun, and Yanning Zhang.
\newblock Attention-based network for low-light image enhancement.
\newblock In \emph{2020 IEEE International Conference on Multimedia and Expo
  (ICME)}, pages 1--6. IEEE, 2020{\natexlab{a}}.

\bibitem[Zhang et~al.(2019{\natexlab{a}})Zhang, Goodfellow, Metaxas, and
  Odena]{zhang2019self}
Han Zhang, Ian Goodfellow, Dimitris Metaxas, and Augustus Odena.
\newblock Self-attention generative adversarial networks.
\newblock In \emph{International Conference on Machine Learning}, pages
  7354--7363, 2019{\natexlab{a}}.

\bibitem[Zhang et~al.(2018{\natexlab{a}})Zhang, Isola, Efros, Shechtman, and
  Wang]{zhang2018unreasonable}
Richard Zhang, Phillip Isola, Alexei~A Efros, Eli Shechtman, and Oliver Wang.
\newblock The unreasonable effectiveness of deep features as a perceptual
  metric.
\newblock In \emph{Proceedings of the IEEE conference on computer vision and
  pattern recognition}, pages 586--595, 2018{\natexlab{a}}.

\bibitem[Zhang et~al.(2018{\natexlab{b}})Zhang, Zhou, Lin, and
  Sun]{zhang2018shufflenet}
Xiangyu Zhang, Xinyu Zhou, Mengxiao Lin, and Jian Sun.
\newblock Shufflenet: An extremely efficient convolutional neural network for
  mobile devices.
\newblock In \emph{Proceedings of the IEEE conference on computer vision and
  pattern recognition}, pages 6848--6856, 2018{\natexlab{b}}.

\bibitem[Zhang et~al.(2019{\natexlab{b}})Zhang, Zhang, and
  Guo]{zhang2019kindling}
Yonghua Zhang, Jiawan Zhang, and Xiaojie Guo.
\newblock Kindling the darkness: A practical low-light image enhancer.
\newblock In \emph{Proceedings of the 27th ACM International Conference on
  Multimedia}, MM '19, pages 1632--1640, New York, NY, USA, 2019{\natexlab{b}}.
  ACM.
\newblock ISBN 978-1-4503-6889-6.
\newblock \doi{10.1145/3343031.3350926}.
\newblock URL \url{http://doi.acm.org/10.1145/3343031.3350926}.

\bibitem[Zhang et~al.(2020{\natexlab{b}})Zhang, Di, Zhang, and
  Wang]{zhang2020self}
Yu~Zhang, Xiaoguang Di, Bin Zhang, and Chunhui Wang.
\newblock Self-supervised image enhancement network: Training with low light
  images only.
\newblock \emph{arXiv}, pages arXiv--2002, 2020{\natexlab{b}}.

\bibitem[Zhang et~al.(2018{\natexlab{c}})Zhang, Tian, Kong, Zhong, and
  Fu]{zhang2018residual}
Yulun Zhang, Yapeng Tian, Yu~Kong, Bineng Zhong, and Yun Fu.
\newblock Residual dense network for image super-resolution.
\newblock In \emph{Proceedings of the IEEE conference on computer vision and
  pattern recognition}, pages 2472--2481, 2018{\natexlab{c}}.

\bibitem[Zhao et~al.(2019)Zhao, Po, Zhang, Liao, Shi, et~al.]{zhao2019saliency}
Yuzhi Zhao, Lai-Man Po, Tiantian Zhang, Zongbang Liao, Xiang Shi, et~al.
\newblock Saliency map-aided generative adversarial network for raw to rgb
  mapping.
\newblock In \emph{2019 IEEE/CVF International Conference on Computer Vision
  Workshop (ICCVW)}, pages 3449--3457. IEEE, 2019.

\bibitem[Zhu et~al.(2020)Zhu, Wang, Jiang, Zong, Liu, Li, and
  Sun]{zhu2020autoassign}
Benjin Zhu, Jianfeng Wang, Zhengkai Jiang, Fuhang Zong, Songtao Liu, Zeming Li,
  and Jian Sun.
\newblock Autoassign: Differentiable label assignment for dense object
  detection.
\newblock \emph{arXiv preprint arXiv:2007.03496}, 2020.

\bibitem[Zhu et~al.(2017)Zhu, Park, Isola, and Efros]{zhu2017unpaired}
Jun-Yan Zhu, Taesung Park, Phillip Isola, and Alexei~A Efros.
\newblock Unpaired image-to-image translation using cycle-consistent
  adversarial networks.
\newblock In \emph{Proceedings of the IEEE international conference on computer
  vision}, pages 2223--2232, 2017.

\bibitem[Zhu et~al.(2021)Zhu, Su, Lu, Li, Wang, and Dai]{zhu2021deformable}
Xizhou Zhu, Weijie Su, Lewei Lu, Bin Li, Xiaogang Wang, and Jifeng Dai.
\newblock Deformable detr: Deformable transformers for end-to-end object
  detection.
\newblock In \emph{International Conference on Learning Representations}, 2021.
\newblock URL \url{https://openreview.net/forum?id=gZ9hCDWe6ke}.

\bibitem[Zhu et~al.(2019)Zhu, Xu, Bai, Huang, and Bai]{annn}
Zhen Zhu, Mengde Xu, Song Bai, Tengteng Huang, and Xiang Bai.
\newblock Asymmetric non-local neural networks for semantic segmentation.
\newblock In \emph{International Conference on Computer Vision}, 2019.
\newblock URL \url{http://arxiv.org/abs/1908.07678}.

\end{thebibliography}
\end{document}